\documentclass[12pt]{article}

\usepackage{setspace}
\usepackage[parfill]{parskip}
\usepackage{paralist}
\usepackage{amsmath}
\usepackage{amssymb}
\usepackage{graphicx}
\usepackage{enumerate}

\usepackage{subcaption} 
\usepackage[super]{nth}
\usepackage{bm}
\usepackage{array}
\usepackage{natbib}
\usepackage[colorlinks,citecolor=blue,urlcolor=blue]{hyperref}
\usepackage{xcolor}
\usepackage{multirow}
\usepackage{enumitem}
\usepackage{algorithm,setspace}
\usepackage{makecell}
\usepackage[noend]{algpseudocode}
\algrenewcommand\algorithmicrequire{\textbf{Input:}}
\algrenewcommand\algorithmicensure{\textbf{Output:}}
\usepackage{mathbbol}
\usepackage{url} % not crucial - just used below for the URL 
\usepackage{theorem}
\newcounter{theorem}
\DeclareMathOperator*{\argmin}{argmin}
\newcommand{\vertiii}[1]{{\left\vert\kern-0.25ex\left\vert\kern-0.25ex\left\vert #1 
    \right\vert\kern-0.25ex\right\vert\kern-0.25ex\right\vert}}
\newcommand{\vertii}[1]{{\left\vert\kern-0.25ex\left\vert #1 
    \right\vert\kern-0.25ex\right\vert}}

\newtheorem{theorem}{Theorem}%[theorem]
\newtheorem{proposition}{Proposition}
\newtheorem{assumption}{Assumption}%[section]
\newtheorem{definition}{Definition}%[section]
\newtheorem{example}{Example}%[section]
\newtheorem{remark}{Remark}%[section]
\newcommand{\blind}{1}

\usepackage{cleveref}
\begin{document}

\def\spacingset#1{\renewcommand{\baselinestretch}%
{#1}\small\normalsize} \spacingset{1}

%%%%%%%%%%%%%%%%%%%%%%%%%%%%%%%%%%%%%%%%%%%%%%%%%%%%%%%%%%%%%%%%%%%%%%%%%%%%%%

\if1\blind
{
  \title{\bf Exponential Family Attention}
  \author{Kevin Christian Wibisono\hspace{.2cm}\\
    Department of Statistics, University of Michigan\\
    and \\
    Yixin Wang \\
    Department of Statistics, University of Michigan}
  \maketitle
} \fi

\if0\blind
{
  \bigskip
  \bigskip
  \bigskip
  \begin{center}
  {\LARGE\bf Exponential Family Attention}
    % {\LARGE\bf Attention-Enhanced Exponential Family Embeddings}
\end{center}
  \medskip
} \fi

\bigskip

% !TeX root = ./jasa-attention.tex

\begin{abstract}

The self-attention mechanism is the backbone of the transformer neural network underlying most large language models. It can capture complex word patterns and long-range dependencies in natural language. This paper introduces \textit{exponential family attention (EFA)}, a probabilistic generative model that extends self-attention to handle high-dimensional sequence, spatial, or spatial-temporal data of mixed data types, including both discrete and continuous observations. The key idea of EFA is to model each observation conditional on all other existing observations, called the context, whose relevance is learned in a data-driven way via an attention-based latent factor model. In particular, unlike static latent embeddings, EFA uses the self-attention mechanism to capture dynamic interactions in the context, where the relevance of each context observations depends on other observations. We establish an identifiability result and provide a generalization guarantee on excess loss for EFA. Across real-world and synthetic data sets---including U.S. city temperatures, Instacart shopping baskets, and MovieLens ratings---we find that EFA consistently outperforms existing models in capturing complex latent structures and reconstructing held-out data.\footnote{Software that replicates the empirical studies can be found at
\url{https://github.com/yixinw-lab/EFA}.}

\end{abstract}

\noindent%

\spacingset{1.2} % DON'T change the spacing!
% !TeX root = ./jasa-attention.tex

\section{Introduction}

The self-attention mechanism and transformers are revolutionary neural network models for processing sequential data~\citep{vaswani2017attention,parikh2016decomposable,lin2017structured,devlin2019bert,brown2020language}. A transformer model builds representations of sequences through multiple layers of self-attention; these representations capture both local and long-range dependencies between word tokens and can model complex linguistic patterns at multiple levels of abstraction. Trained transformer models have dramatically advanced our ability to process language and have enabled breakthrough performance across a wide range of language understanding and generation tasks, from machine translation to question answering to text summarization.

\begin{figure}[t]
\centering
\includegraphics[width=16.25cm]{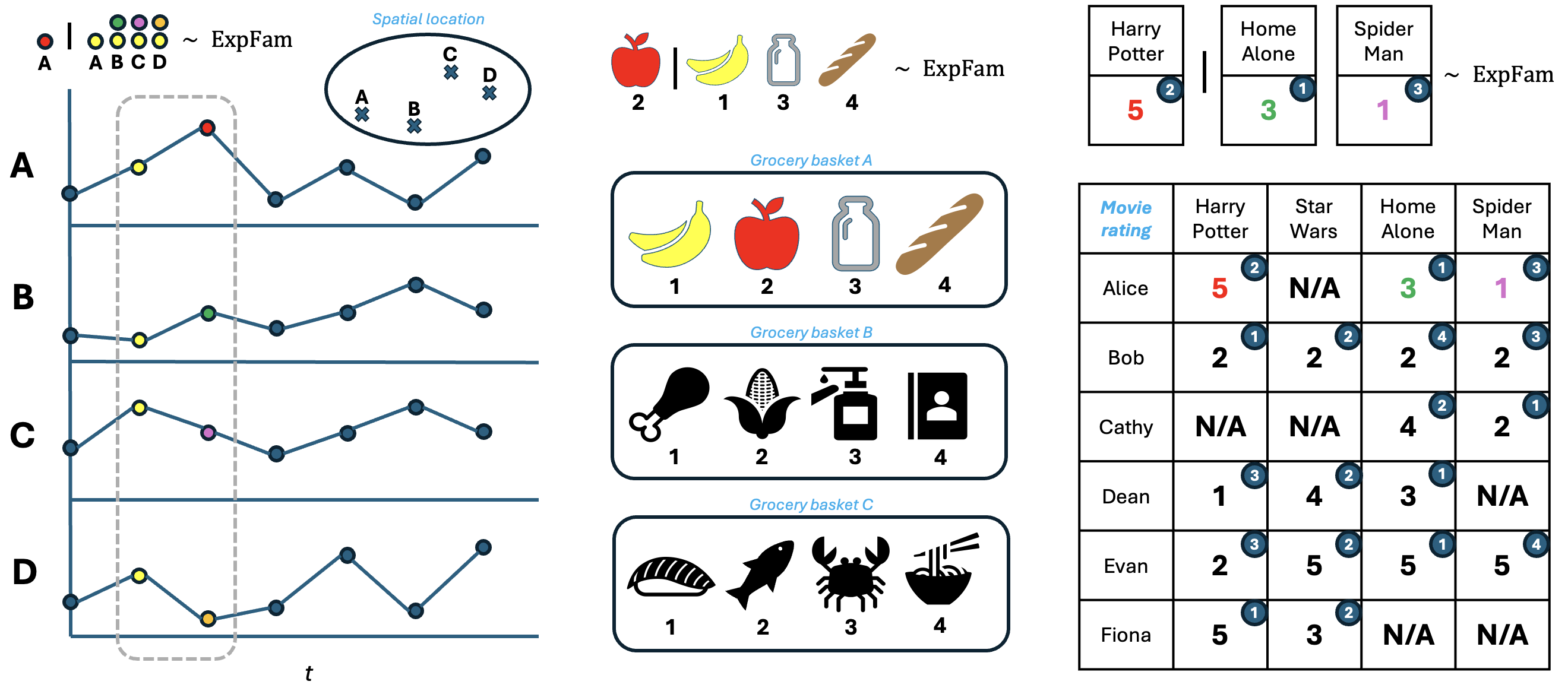}
\caption{Examples of data that can be effectively modeled by exponential family attention (EFA) include: (\textit{left}) spatiotemporal data, where each time series is linked to a specific attribute (e.g., spatial location); (\textit{center}) sequential grocery purchase data, capturing the sequence of purchased items and their quantities for each user; and (\textit{right}) sequential movie rating data, detailing the order of movies rated by each user along with their corresponding ratings.}
\label{fig:efa}
\end{figure}

The self-attention mechanism models each word (token) based on its context, comprising all preceding words (tokens). The likelihood of a word given its context is determined using a latent factor model, akin to matrix factorization. Specifically, this likelihood is proportional to a weighted sum of the inner products between the latent embedding of this word and the embeddings of its context words (known as \textit{value} vectors). The weights, representing the relevance of each context word, are learned in a data-driven manner through another latent factor model. In this model, the relevance of a context word to the current word is determined by the inner product of another set of embeddings: the \textit{query} vector for the current word and the \textit{key} vector for the context word. Self-attention dynamically captures context-dependent interactions, where the relevance of each context word depends on others. By training on a text corpus, the model learns parameters and embeddings by maximizing the likelihood of predicting each word given its context, effectively identifying which context words are most relevant for prediction.

\noindent
\textbf{Exponential family attention.} This paper introduces exponential family attention (EFA), a probabilistic generative model that extends self-attention beyond natural language to high-dimensional sequence, spatial, or spatiotemporal data with mixed data types, including both discrete and continuous observations (see Figure \ref{fig:efa}). Our motivation is that many other types of data can benefit from the same intuition that underlies self-attention: the occurrence of elements should influence the likelihood of others based on their relevance and compatibility, both of which can be learned from the data. To generalize self-attention to other data types, we leverage exponential families to extend the self-attention mechanism, letting the scaled dot-product attention weights inform the natural parameters of exponential family distributions. This approach enables the modeling of diverse dependencies and interactions between elements while maintaining the flexibility of the traditional self-attention mechanism.

\noindent
\textbf{Applications of EFA.} As one example beyond natural language data, we can study spatiotemporal temperature data across U.S. cities (see left of Figure \ref{fig:efa}). Meteorologists measure temperature changes across many locations over time, seeking to understand the complex patterns of weather systems and climate dynamics. In this setting, self-attention can model how each city's temperature relates to temperatures observed in other locations, with attention weights capturing both temporal and spatial dependencies. The temperature from a target city is driven by earlier temperatures from other cities, where the relevance (attention) weights are flexible learned functions of both temporal offsets and spatial distances. For example, the model might learn that a city's temperature is strongly influenced by conditions in cities to its west from several hours prior, while having weaker dependencies on cities in other directions or at longer time delays.

Another example we study involves users rating movies on streaming platforms (see right of Figure \ref{fig:efa}). Streaming services collect sequential rating data and are interested in understanding and predicting user behavior to improve recommendations and user experience. In this setting, self-attention can model how each rating action depends on the context of previous ratings, where the attention mechanism learns complex patterns across multiple types of information. When predicting the next rated movie and its rating, the likelihood of a candidate movie-rating pair is driven by the latent representations of all previous rating actions, with the relevance (attention) weights capturing dependencies on the movie features (like genre, director, actors), temporal patterns (such as seasonal preferences or rating frequency), and previous rating values. This attention-based framework can capture fine-grained behavioral patterns, such as how users binge-rate similar movies in short time spans, explore new genres after particularly high or low ratings, or show periodic preferences for different content. The model learns these patterns while naturally handling the mixed categorical (movie identity, genre) and continuous (timestamps, ratings) data.

\noindent
\textbf{Ingredients of EFA.} EFA has three ingredients. (1) For each observation, we specify \textit{three types of embeddings}---query, key, and value vectors---that encode different aspects of the data. For example, in climate data, these vectors might encode location coordinates, temporal features, and weather conditions. (2) We define the appropriate \textit{conditional exponential families} for different types of observations, such as Gaussian distributions for continuous temperature measurements and ordinal distributions for ratings. (3) We specify the \textit{attention mechanism} that characterizes how the embeddings interact in modeling dependencies, e.g., a query-key inner product mapped through a non-linear function that respects the data types involved. These three ingredients enable us to construct flexible attention models that respect the statistical nature of diverse data types while capturing complex dependencies.

\noindent
\textbf{This work.} We develop the EFA model and establish its theoretical identifiability as well as generalization guarantees. We also show how variants of latent factor models can be viewed as special cases of EFA with specific attention structures and parameter-sharing patterns. This connection provides new insights into both the expressiveness of attention mechanisms and the limitations of latent factor modeling. We evaluate EFA across diverse real-world data sets---including U.S. city temperatures, Instacart shopping baskets, and MovieLens ratings---where it consistently outperforms traditional latent factor models in capturing complex latent structures and reconstructing held-out data. Our work builds upon the related works from latent factor models, exponential family embeddings, and the self-attention mechanism, detailed in \Cref{sec:related}.

% \input{related}
% !TeX root = ./jasa-attention.tex

\section{Exponential family attention}
\label{sec:exp-fam-attention}
In this section, we begin by reviewing the self-attention mechanism as is typically employed in modeling natural language data (Section \ref{sec:self-attn-lang}). We then introduce exponential family attention (EFA), a class of probabilistic models that extends the self-attention mechanism to model both sequential and non-sequential data beyond text (Section \ref{sec:self-attn-non-lang}).

\subsection{Self-attention for modeling natural language data}
\label{sec:self-attn-lang}

In self-attention, each word is associated with three learned embeddings---query, key, and value vectors---that govern how it interacts with and aggregates information from preceding words (also known as context words). The conditional probability of the next word is computed by taking scaled dot products between its query vector and the key vectors of preceding words, followed by a softmax operation. This produces attention weights that determine the relevance of each preceding word. These weights are then used to aggregate the value vectors of the preceding words, which informs the likelihood of the next word. During training, the query, key, and value vectors are optimized by maximizing the likelihood of predicting each word given its context. This enables the model to learn, in a data-driven manner, which context words are most relevant for predicting the next word.

Formally, consider sentences of the form $x = x_{1:I}$, where $I$ denotes the sentence length. Also, let $D$ be the vocabulary size (thus $x_i \in [D]$ for each $i \in [I]$). Suppose that our goal is to model these sentences in a unidirectional manner. That is, we would like to estimate some parameter $\bm{\theta}$ that maximizes the log-likelihood, i.e., $\log p(x; \bm{\theta}) := \sum_{i \in [I]} \log p(x_i \mid x_{1:i-1}; \bm{\theta})$, across all sentences. 

In a standard self-attention model, each token $x_i$ is distributed as $$p(x_i \mid x_{1:i-1}; \bm{\theta}) \sim \textrm{Categorical}(\eta_{i; \bm{\theta}}(x_{1:i-1})),$$
where $\bm{\theta}$ denotes the set of all parameters; the probability vector $\eta_{i; \bm{\theta}}(x_{1:i-1})$ is derived as follows.
\begin{enumerate}[leftmargin=0.6cm]
    \item Mask the $i$-th token, i.e., the sentence becomes $(x_1, \dots, x_{i-1}, \texttt{[MASK]}, x_{i+1}, \dots, x_I)$.
    \item Add the context and positional embeddings of the masked sentence:
    $$X_i := \left[ \bm{\beta}_{x_1} + \bm{p}_1, \cdots, \bm{\beta}_{x_{i-1}} + \bm{p}_{i-1}, \bm{\beta}_{\texttt{[MASK]}} + \bm{p}_i, \bm{\beta}_{x_{i+1}} + \bm{p}_{i+1}, \cdots, \bm{\beta}_{x_{I}} + \bm{p}_{I} \right] \in \mathbb{R}^{K \times I},$$
    % \vspace{-3mm}
    where $\bm{\beta}_\ell \in \mathbb{R}^K$ is the  \textit{context embedding} for each token $\ell \in [D]$  as well as the \texttt{[MASK]} token, compactly represented as $\bm{\beta} \in \mathbb{R}^{K \times (D+1)}$, and $\bm{p}_i \in \mathbb{R}^K$ for each position $i \in [I]$ is the \textit{positional embedding} for each, compactly represented as $\bm{P} \in \mathbb{R}^{K \times I}$.
    \item Transform $X_i$ via self-attention, where the softmax is applied column-wise:
    \begin{align}
    \label{eq:attention}
    X_i' := \bm{W}^\mathcal{V}X_i \cdot \textrm{softmax}\left(\frac{\left( \bm{W}^\mathcal{Q} X_i \right)^\top \bm{W}^\mathcal{K} X_i}{K} + M \right) \in \mathbb{R}^{K \times I},
    \end{align}
    where $\bm{W}^\mathcal{Q}, \bm{W}^\mathcal{K}, \bm{W}^\mathcal{V} \in \mathbb{R}^{K \times K}$ are the \textit{query}, \textit{key} and \textit{value matrices}  that transforms the context and positional embeddings of each word into their query, key, and value vectors; the mask 
    $M \in \mathbb{R}^{I \times I}$ satisfies that $M_{\star, \bm{\cdot}} = 0$ if $\star \leq \bm{\cdot}$ and $M_{\star, \bm{\cdot}} = -\infty$ otherwise, a.k.a. a \textit{causal mask}.\footnote{The word ``causal'' in ``causal mask'' is unrelated to causal inference.} 
    \item Define 
    $\eta_{i; \bm{\theta}}(x_{1:i-1}) := \sigma\left( \bm{\delta}^\top X_{ii}' \right) \in \mathbb{R}^D$,
    where $\bm{\delta}_\ell \in \mathbb{R}^K$ is \textit{center embedding}  for each token $\ell \in [D]$, compactly represented as $\bm{\delta} \in \mathbb{R}^{K \times D}$; the $\sigma$ and $X_{ii}'$ denote the softmax operator and the $i$-th column of $X_i'$, respectively.
\end{enumerate}

\begin{remark}[Self-attention and latent factor modeling.] Steps 3 and 4 (especially Eq-uation \ref{eq:attention}) tie the self-attention mechanism to latent factor modeling. They build on a latent factor model structure $\sigma\left( \bm{\delta}^\top (\bm{W}^\mathcal{V}X_i)_{ii}' \right)$, further reweighted by attention weights---derived from another latent factor model based on the learned query and key matrices $\textrm{softmax}\left(\left( \bm{W}^\mathcal{Q} X_i \right)^\top \bm{W}^\mathcal{K} X_i\right)$ with additional scaling (by $K$) and shifting (by $M$). This learned factor model for attention weights allow each word preceding the masked word to interact flexibly when predicting the masked word.
\end{remark}

% \end{enumerate}
% \begin{enumerate}[leftmargin=0.6cm]
% \item A \textit{center embedding} $\bm{\delta}_\ell \in \mathbb{R}^K$ for each token $\ell \in [D]$, compactly represented as $\bm{\delta} \in \mathbb{R}^{K \times D}$;
% \item A \textit{context embedding} $\bm{\beta}_\ell \in \mathbb{R}^K$ for each token $\ell \in [D]$ as well as the \texttt{[MASK]} token, compactly represented as $\bm{\beta} \in \mathbb{R}^{K \times (D+1)}$;
% \item A \textit{positional embedding} $\bm{p}_i \in \mathbb{R}^K$ for each position $i \in [I]$, compactly represented as $\bm{P} \in \mathbb{R}^{K \times I}$; and
% \item \textit{Query}, \textit{key} and \textit{value matrices} $\bm{W}^\mathcal{Q}, \bm{W}^\mathcal{K}, \bm{W}^\mathcal{V} \in \mathbb{R}^{K \times K}$ that transforms the context and positional embeddings of each word into their query, key, and value vectors.
% \end{enumerate}

\begin{remark}[Extensions to multi-head multi-layer self-attention model]
    In practi-ce, additional parameters may arise from the use of multi-head attention, multi-layer attention, feed-forward layers, and layer normalization (e.g., \citet{vaswani2017attention,devlin2019bert}). We omit these aspects for ease of exposition. All the discussions remain valid with these additional components.
\end{remark}

\begin{remark}[Causal mask ensures unidirectionality]
    The causal mask in Step 3 ensures that the probability vector for next token prediction $\eta_{i; \bm{\theta}}(x_{1:i-1})$ does not depend on any future token $x_{i+1}, \dots, x_I$, making the model \emph{unidirectional} (right depends on left), as opposed to \emph{bidirectional}.
\end{remark}

\noindent
\textbf{Estimation of parameters.} Given training text sequences $\{x^f\}_{f=1}^F$, where $x^f = x^f_{1:I}$ for each $f \in [F]$, the parameters $\bm{\theta}$ is estimated by maximizing the log-likelihood 
$$\ell \left(\{x^f\}_{f=1}^F \mid \bm{\theta} \right) := \frac{1}{FI} \sum_{f=1}^F \sum_{i=1}^I \left(\log \eta_{i; \bm{\theta}}(x^f_{1:i-1})\right)_{x^f_i},$$
where $(\bm{\cdot})_\star$ represents the $\star$-th entry of the vector $\bm{\cdot}$.

\begin{remark}[Extension to bidirectional modeling] The model $p(x_i \mid x_{1:i-1}; \bm{\theta})$ above can be adapted for bidirectional modeling  $p(x_i \mid x_{1:i-1}, x_{i+1:I}; \bm{\theta})$ by setting the causal mask $M$ to zero and optimizing the pseudo log-likelihood.
\end{remark}

\subsection{EFA: Extending self-attention beyond text data}
\label{sec:self-attn-non-lang}
In Section \ref{sec:self-attn-lang}, we described how the self-attention mechanism can be used to model text data. We next extend the self-attention mechanism beyond text. We start with a market basket example.

\begin{example}[Market baskets; \citet{chen2020studying}]
\label{ex:baskets}
    In market basket analysis, we are given $J$ market baskets, each containing an ordered list of items purchased together in a single transaction, including both the sequence of items purchased and the quantity of each item.
    % We can treat these baskets as text, where items and baskets correspond to words and sentences, respectively. When the baskets are unordered, a bidirectional model with no positional embeddings can be used.
\end{example}

Like in Example \ref{ex:baskets}, many sequential datasets we analyze extend beyond simple sequences of categorical observations like text data. Rather, each observation is often associated with not only categorical observations but also an associated value that can provide additional information. For example, the market basket analysis involves both the categorical sequence of items purchased and the quantity of each. Similarly, in movie rating analysis, we might have access to not only the sequence of movies rated by a user, but also the rating that the user assigns to each movie. To effectively and flexibly model such data, we propose \emph{exponential family attention (EFA).}

\begin{figure}[t]
\centering
\includegraphics[width=16.25cm]{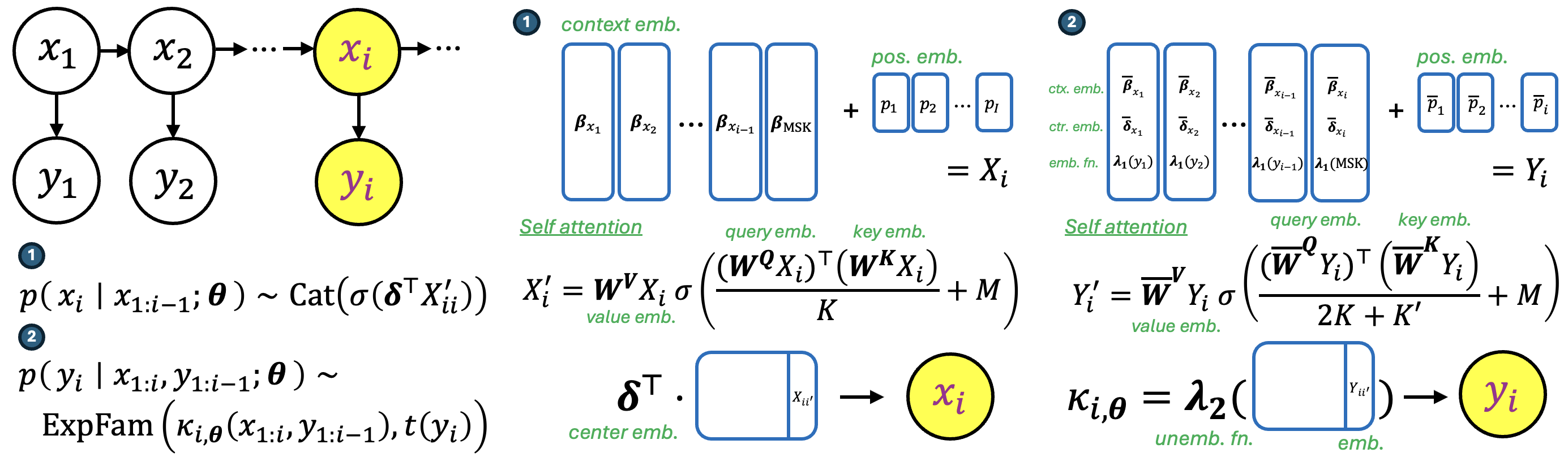}
\caption{An end-to-end illustration of the exponential family attention (EFA) model. The middle and right panels correspond to the first and second terms of Equation \eqref{eq:decomp}, respectively. Here, $\sigma(\cdot)$ and $\textrm{Cat}(\cdot)$ refer to the softmax operation and categorical distribution, respectively.}
\label{fig:efa-ilust}
\end{figure}

\noindent
\textbf{The exponential family attention (EFA) model.} Consider data points of the form $(x, y) = (x_{1:I}, y_{1:I})$, where $x_i$'s are categorical with $D$ categories (as in Section \ref{sec:self-attn-lang}), and $y_i$'s ($\in \mathbb{R}^{d_y}$) follow an exponential family distribution (e.g., Gaussian or Poisson). EFA models the data in a unidirectional sequential manner, assuming the following decomposition:
\begin{equation}
\label{eq:decomp}
   \log p(x,y; \bm{\theta}) := \sum_{i \in [I]} \left(\log p(x_i \mid x_{1:i-1}; \bm{\theta}) + \log p(y_i \mid x_{1:i}, y_{1:i-1}; \bm{\theta}) \right),
\end{equation}
that is, assuming $x_i \perp y_{1:i-1} \mid x_{1:i-1}$. We model the categorical component $p(x_i \mid x_{1:i-1}; \bm{\theta})$ using the self-attention structure in Section \ref{sec:self-attn-lang}. We then introduce an exponential family module to model the values associated with the categories:
$$p(y_i \mid x_{1:i}, y_{1:i-1}; \bm{\theta}) \sim \textrm{ExpFam}(\kappa_{i; \bm{\theta}}(x_{1:i}, y_{1:i-1}), t(y_i)),$$ where $t(y_i)$ is the sufficient statistic and $\kappa_{i; \bm{\theta}}(x_{1:i}, y_{1:i-1})$ is the natural parameter derived below:
\begin{enumerate}[leftmargin=0.6cm]
    \item Concatenate the context embeddings ($\bm{\bar{\delta}}_\ell \in \mathbb{R}^K$ for each token $\ell \in [D]$, compactly represented as $\bm{\bar{\delta}})$ and center embeddings ($\bm{\bar{\beta}}_\ell \in \mathbb{R}^K$ for each token $\ell \in [D]$, compactly represented as $\bm{\bar{\beta}}  \in \mathbb{R}^{K \times D}$) of the $x_i$'s, along with applying an embedding function $\bm{\lambda_1}(\cdot): \mathbb{R}^{d_y}\rightarrow \mathbb{R}^{K'}$ to each $y_{1:I}$ (with $y_i$ being masked):
    \[
    Y_i := 
    \left[
      \begin{array}{ccccccc}
        \bm{\bar{\beta}}_{x_1} & \ldots & \bm{\bar{\beta}}_{x_{i-1}} & \bm{\bar{\beta}}_{x_i} & \bm{\bar{\beta}}_{x_{i+1}} & \ldots & \bm{\bar{\beta}}_{x_{I}} \\
        \bm{\bar{\delta}}_{x_1} & \ldots & \bm{\bar{\delta}}_{x_{i-1}} & \bm{\bar{\delta}}_{x_i} & \bm{\bar{\delta}}_{x_{i+1}} & \ldots & \bm{\bar{\delta}}_{x_{I}} \\
        \bm{\lambda_1}(y_1) & \ldots & \bm{\lambda_1}(y_{i-1}) & \bm{\lambda_1}(\texttt{[MASK]}) & \bm{\lambda_1}(y_{i+1}) & \ldots & \bm{\lambda_1}(y_{I})
      \end{array}
    \right] \in \mathbb{R}^{(2K+K') \times I},
\]
    \item Add positional embeddings ($\bm{\bar{p}}_i \in \mathbb{R}^{2K+K'}$ for each position $i \in [I]$, compactly represented as $\bm{\bar{P}} \in \mathbb{R}^{(2K+K') \times I}$): $Y_i := Y_i + \bm{\bar{P}} \in \mathbb{R}^{(2K+K') \times I}$. 
    \item Transform $Y_i$ via self-attention, where the softmax is applied column-wise:
    $$Y_i' := \bm{\bar{W}}^\mathcal{V}Y_i \cdot \textrm{softmax}\left(\frac{\left( \bm{\bar{W}}^\mathcal{Q} Y_i \right)^\top \bm{\bar{W}}^\mathcal{K} Y_i}{2K + K'} + M \right) \in \mathbb{R}^{(2K + K') \times I},$$
    where $\bm{\bar{W}}^\mathcal{Q}, \bm{\bar{W}}^\mathcal{K}, \bm{\bar{W}}^\mathcal{V} \in \mathbb{R}^{(2K+K') \times (2K+K')}$  are \textit{query}, \textit{key} and \textit{value matrices}, and $M \in \mathbb{R}^{I \times I}$ is a \textit{causal mask} such that $M_{\star, \bm{\cdot}} = 0$ if $\star \leq \bm{\cdot}$ and $M_{\star, \bm{\cdot}} = -\infty$ otherwise.
    \item Define 
    $\kappa_{i; \bm{\theta}}(x_{1:i}, y_{1:i-1}) := \bm{\lambda_2}(Y_{ii}') \in \mathbb{R}^{d_y}$,
    where $Y_{ii}'$ denotes the $i$-th column of $Y_i'$.
\end{enumerate}
Refer to Figure \ref{fig:efa-ilust} for a complete illustration of the EFA model.

\begin{remark}[Uni-/Bi-directional modeling]
    Similar to Section \ref{sec:self-attn-lang}, the causal mask $M$ ensures that no information from future tokens affects $\kappa_{i; \bm{\theta}}(x_{1:i}, y_{1:i-1})$, i.e. unidirectional modeling. It can be modified to bidirectional modeling by setting $M=\boldsymbol{0}$ and optimizing the pseudo log-likelihood.
\end{remark}
\begin{remark}[Choices of embedding and unembedding functions]
    Many choices exist for the embedding and unembedding functions. For instance, one can define $\bm{\lambda_1}(\cdot)$ as the identity function with value zero for the \texttt{[MASK]} token, and $\bm{\lambda_2}(\cdot)$ as reading off the last $d_y$ entries of a vector. Alternatively, when the $y_i$'s only consist of a few possible values, we can define $\bm{\lambda_1}(\cdot)$ to map each possible value to a learned embedding, along with a separate embedding for the \texttt{[MASK]} token.
\end{remark}
\begin{remark}[Center and context embeddings]
    We introduced center and context embeddings, $\bm{\bar{\delta}}$ and $\bm{\bar{\beta}}$, for consistency with Section \ref{sec:self-attn-lang}. Using just one of them is adequate in practice.
\end{remark}

\begin{remark}[Extensions to multi-head multi-layer exponential family attention]
     One can utilize multi-head and multi-layer attention, feed-forward layers, layer normalization layers, and residual connections to increase the expressivity of the EFA model. For instance, with multi-layer attention, we can repeat Step 3 multiple (e.g., $L > 1$) times with separate query, key and value matrices $\bm{\bar{W}}^{\mathcal{Q}^{(l)}}, \bm{\bar{W}}^{\mathcal{K}^{(l)}}, \bm{\bar{W}}^{\mathcal{V}^{(l)}}$ for each $l \in [L]$. We can then recursively define
    $$Y_i^{(l)} := \bm{\bar{W}}^{\mathcal{V}^{(l)}}Y_i^{(l-1)} \cdot \textrm{softmax}\left(\frac{\left( \bm{\bar{W}}^{\mathcal{Q}^{(l)}} Y_i^{(l-1)} \right)^\top \bm{\bar{W}}^{\mathcal{K}^{(l)}} Y_i^{(l-1)}}{2K + K'} + M \right) \in \mathbb{R}^{(2K + K') \times I}$$
    where $Y_i^{(0)} := Y_i$, and finally set $Y_i' := Y_i^{(L)}$.
    
\end{remark}
\textbf{Estimation of parameters.} Given training data $\{(x,y)^f\}_{f=1}^F$, where $(x,y)^f = (x^f_{1:I},y^f_{1:I})$ for each $f \in [F]$, the parameters $\bm{\theta}$ can be estimated by maximizing the following log-likelihood:
\begin{align}
\label{eq:ll}
    &\ell \left(\{(x,y)^f\}_{f=1}^F \mid \bm{\theta} \right) \notag \\
    &:= \frac{1}{FI} \sum_{f=1}^F \sum_{i=1}^I \left(\left(\log \eta_{i; \bm{\theta}}(x^f_{1:i-1})\right)_{x^f_i} + \left(\kappa_{i; \bm{\theta}}(x^f_{1:i}, y^f_{1:i-1})\right)^\top t(y_i^f) - A\left( \kappa_{i; \bm{\theta}}(x^f_{1:i}, y^f_{1:i-1}) \right) \right),
\end{align}
where $(\bm{\cdot})_\star$ denotes the $\star$-th entry of the vector $\bm{\cdot}$ and $A(\bm{\cdot})$ denotes the log-partition function of the exponential family distribution. Algorithm \ref{alg:est} summarizes the steps needed to estimate the parameters of an EFA model from the training data.

\begin{algorithm}[t]
\setstretch{1.5}
\caption{Estimation of parameters in EFA}
\label{alg:est}
\begin{algorithmic}[1]
\Require Data $(x, y)^f = (x^f_{1:I}, y^f_{1:I})$ for each $f \in [F]$, where the $x^f_i$'s ($\in [D]$) are categorical and the $y^f_i$'s ($\in \mathbb{R}^{d_y}$) follow an exponential family distribution.
\Ensure Estimated parameters $\hat{\bm{\theta}}$
% where 
% $\bm{\theta} := (\bm{\delta}, \bm{\beta}, \bm{P}, \bm{\bar{\delta}}, \bm{\bar{\beta}}, \bm{\bar{P}}, \bm{W}^\mathcal{Q}, \bm{W}^\mathcal{K}, \bm{W}^\mathcal{V}, \bm{\bar{W}}^\mathcal{Q}, \bm{\bar{W}}^\mathcal{K}, \bm{\bar{W}}^\mathcal{V}, \bm{\lambda_1}(\cdot), \bm{\lambda_2}(\cdot)).$

\State Define $\eta_{i; \bm{\theta}}(x^f_{1:i-1})$ for each $f \in [F]$ and $i \in [I]$ following Steps 1 to 4 in Section \ref{sec:self-attn-lang}.
\State Define $\kappa_{i; \bm{\theta}}(x^f_{1:i}, y^f_{1:i-1})$ for each $f \in [F]$ and $i \in [I]$ following Steps 1 to 4 in Section \ref{sec:self-attn-non-lang}.
\State Define the log-likelihood following Equation \eqref{eq:ll}, where $t(\cdot)$ and $A(\cdot)$ denote the sufficient statistic and partition function of the exponential family distribution, respectively.
\State \Return $\hat{\bm{\theta}}=\arg\max \ell \left(\{(x,y)^f\}_{f=1}^F \mid \bm{\theta} \right)$
\end{algorithmic}
\end{algorithm}
% !TeX root = ./jasa-attention.tex

\section{Examples of exponential family attention}
\label{sec:ex-of-efa}
In this section, we present three distinct instantiations of exponential family attention (EFA) for modeling real-world data, namely market baskets (Section \ref{sec:app-baskets}), Gaussian spatiotemporal time series (Section \ref{sec:app-gauss}), and movie ratings (Section \ref{sec:app-movie}).

\subsection{Market baskets}
\label{sec:app-baskets}
We begin by revisiting the market basket scenario discussed in Example \ref{ex:baskets} of Section \ref{sec:self-attn-non-lang}. Suppose there are $D$ possible items and $F$ baskets, where each basket $x^f = x^f_{1:I}$ consists of $I$ distinct items. The assumption of each basket containing the same number of items is made solely for notational convenience. Drawing analogies to text data, we consider each basket as a piece of text, where items and baskets correspond to words and sentences, respectively. 

When the order of items in each basket is known, we can use either the unidirectional or bidirectional version of the method described in Section \ref{sec:self-attn-lang} to learn center and context embeddings (i.e., representations) for each item. If the order is not known, we can apply the bidirectional version of the same method without positional embeddings. In both scenarios, the learned embeddings for each item can help downstream analysis, such as identifying complements and substitutes \citep{chen2020studying} or serving as features for subsequent prediction or classification tasks.

\subsection{Spatiotemporal time series}
\label{sec:app-gauss}
We next consider an example of spatiotemporal time series data, which we use EFA to model.
\begin{example}[Gaussian spatiotemporal time series; 
\citet{rudolph2016exponential}]
\label{ex:gauss}
    Given \\$I$ time series, each of which contains a single observation at each time step from $1$ to $F$. Denote the observation at time step $f$ for the $i$-th time series as $y_{(i,f)} \in \mathbb{R}$. Each time series is associated with an attribute $\tau_i \in \mathbb{R}^\tau$. For example, if the goal is to model the activity of a neuron at a given time conditional on the activity of other neurons at the same time, then each neuron's spatial location can be considered its attribute. 
\end{example}

In this example, we focus on modeling the second part of Equation \eqref{eq:decomp}, as we consider the same $I$ time series at each time step. Additionally, our model must be bidirectional and not contain positional embeddings, as the order in which the time series are observed is irrelevant in this case. Finally, we assume that each time series has a learned embedding $\bm{g}(\tau_i)$ for some (non-linear) function $\bm{g}: \mathbb{R}^\tau \rightarrow \mathbb{R}^K$. This formulation offers more flexibility to generate embeddings for time series that are not in the training data, such as neurons at new spatial locations.

% yw / note for later: i do not fully understand the paragraph above.

We now present an EFA model for Example \ref{ex:gauss}. The goal is to model $y_{(i,f)} \in \mathbb{R}$ conditional on the observations at time step $f$ for the other time series. We omit the time index $f$ for notational convenience.
\begin{enumerate}[leftmargin=0.6cm]
\item Denote
$$Y_i := 
\left[
  \begin{array}{ccccccc}
    \bm{g}(\tau_1) & \ldots & \bm{g}(\tau_{i-1}) & \bm{g}(\tau_i) & \bm{g}(\tau_{i+1}) & \ldots & \bm{g}(\tau_I) \\
    \bm{\lambda_1}(y_1) & \ldots & \bm{\lambda_1}(y_{i-1}) & \bm{\lambda_1}(\texttt{[MASK]}) & \bm{\lambda_1}(y_{i+1}) & \ldots & \bm{\lambda_1}(y_{I})
  \end{array}
\right] \in \mathbb{R}^{(K+K') \times I}$$
for some learned $\bm{\lambda_1}: \mathbb{R} \rightarrow \mathbb{R}^{K'}$. As we are interested in predicting $y_i$, we replace it with the \texttt{[MASK]} token.
\item Introduce \textit{query}, \textit{key} and \textit{value} matrices: $\bm{\bar{W}}^\mathcal{Q}, \bm{\bar{W}}^\mathcal{K}, \bm{\bar{W}}^\mathcal{V} \in \mathbb{R}^{(K+K') \times (K+K')}$.
\item Calculate the attention-transformed matrix, where the softmax is applied column-wise:
    $$Y_i' := \bm{\bar{W}}^\mathcal{V}Y_i \cdot \textrm{softmax}\left(\frac{\left( \bm{\bar{W}}^\mathcal{Q} Y_i \right)^\top \bm{\bar{W}}^\mathcal{K} Y_i}{K + K'} \right) \in \mathbb{R}^{(K + K') \times I}.$$
\item The masked observation $y_i$ satisfies
$y_i \mid y_{-i} \sim \mathcal{N}(\bm{\lambda_2}(Y'_{ii}), \sigma^2)$
for some $\bm{\lambda_2}: \mathbb{R}^{K + K'} \rightarrow \mathbb{R}$ and variance $\sigma^2$. Here, $Y'_{ii}$ represents the $i$-th column of $Y_i'$.
\end{enumerate}

\begin{remark}[Extensions of EFA]
   More complex non-linear operations can be incorporated into EFA, such as multi-head multi-layer attention, feed-forward layers, and residual connections.
\end{remark}

\subsection{Movie ratings}
\label{sec:app-movie}
We next study how EFA can be employed to model movie ratings.
\begin{example}[Movie ratings; \citet{rudolph2016exponential}] 
\label{ex:ratings}
We are given $D$ movies and $F$ users. For each movie $d \in [D]$ and user $f \in [F]$, let $y_{(d,f)} \in \{1, 2, 3\}$ be the rating given by user $f$ for movie $d$, provided that user $f$ has rated movie $d$. For notational convenience, assume that each user has rated $I$ movies. Moreover, for each user $f$, we observe the data $(x^f, y^f) = (x^f_{1:I}, y^f_{1:I})$, where $x^f_{1:I}$ denotes the ordered list of movies rated by user $f$, and $y^f_{1:I}$ denotes the corresponding ratings.% We define $c_i := c_{(n,t)} = \{ (m,t) \mid x_{(m,t)} \hspace{2mm} exists\}$ and model the data as
% $$x_{i=(n,t)} - 1 \mid x_{c_i} \sim \textrm{Poisson}\left( \exp \left(  \rho_n^\top \hspace{1mm} \frac{\sum_{(m,t) \in c_i} \alpha_m x_{(m,t)}}{\left| c_i \right|} \right)\right)$$
% or
% $$x_{i=(n,t)} \mid x_{c_i} \sim \textrm{Poisson}\left( 1 + \exp \left(  \rho_n^\top \hspace{1mm} \frac{\sum_{(m,t) \in c_i} \alpha_m x_{(m,t)}}{\left| c_i \right|} \right)\right).$$
% Note that the center and context embeddings for each movie are the same across all users. 
\end{example}

In order to use EFA in Example \ref{ex:ratings}, we can follow the unidirectional or bidirectional EFA model described in Section \ref{sec:self-attn-non-lang}. Concretely, for each $i \in [I]$, we model $x_i \mid x_{1:i-1}$ and $y_i \mid x_{1:i}, y_{1:i-1}$ in the unidirectional case, or $x_i \mid x_{-i}$ and $y_i \mid x_{1:I}, y_{-i}$ in the bidirectional case. We drop the user index $f$ for notational convenience.

We now present a simplified version of EFA that bidirectionally models the second term of Equation \eqref{eq:decomp} using linear attention instead of softmax attention. Suppose we are interested in modeling the rating $y_i$ conditional on $x_{1:I}$ and $y_{-i}$. The EFA model is as follows:
\begin{enumerate}[leftmargin=0.6cm]
    \item Let
    $Y_i := 
    \left[
      \begin{array}{ccccccc}
        \bm{\bar{\beta}}_{x_1} & \ldots & \bm{\bar{\beta}}_{x_{i-1}} & \bm{\bar{\beta}}_{x_i} & \bm{\bar{\beta}}_{x_{i+1}} & \ldots & \bm{\bar{\beta}}_{x_{I}} \\
        \bm{\bar{\delta}}_{x_1} & \ldots & \bm{\bar{\delta}}_{x_{i-1}} & \bm{\bar{\delta}}_{x_i} & \bm{\bar{\delta}}_{x_{i+1}} & \ldots & \bm{\bar{\delta}}_{x_{I}} \\
        \bm{\lambda_1}(y_1) & \ldots & \bm{\lambda_1}(y_{i-1}) & \bm{\lambda_1}(\texttt{[MASK]}) & \bm{\lambda_1}(y_{i+1}) & \ldots & \bm{\lambda_1}(y_{I})
      \end{array}
    \right] \in \mathbb{R}^{(2K+K') \times I}$

for some learned $\bm{\lambda_1}: \mathbb{R} \rightarrow \mathbb{R}^{K'}$. As we are interested in predicting $y_i$, we replace it with the the \texttt{[MASK]} token.

    \item Add positional embeddings: $Y_i := Y_i + \bm{\bar{P}} \in \mathbb{R}^{(2K+K') \times I}$. 
    \item Transform $Y_i$ via self-attention, where the softmax is applied column-wise:
    $$Y_i' := \bm{\bar{W}}^\mathcal{V}Y_i \cdot \left(\frac{\left( \bm{\bar{W}}^\mathcal{Q} Y_i \right)^\top \bm{\bar{W}}^\mathcal{K} Y_i}{2K + K'} \right) \in \mathbb{R}^{(2K + K') \times I}.$$
    \item The masked observation $y_i$ satisfies
$y_i -1 \mid y_{-i} \sim \textrm{Poisson}(\exp(\bm{\lambda_2}(Y'_{ii})))$
for some $\bm{\lambda_2}: \mathbb{R}^{2K + K'} \rightarrow \mathbb{R}$. Here, $Y'_{ii}$ represents the $i$-th column of $Y_i'$.
\end{enumerate}

\begin{remark}[Extensions of EFA]
    More complex non-linear operations can be incorporated into EFA, such as multi-head multi-layer attention, feed-forward layers, and residual connections.
\end{remark}

\begin{remark}[Extensions to other types of ratings data]
    The EFA formulation directly extends to the case where the ratings $y_i$'s are Bernoulli or Gaussian (e.g., see Section \ref{sec:synth}).
\end{remark}

% !TeX root = ./jasa-attention.tex

\section{Exponential family attention as an extension of latent factor models}
\label{sec:conn}

In this section, we compare and constrast exponential family attention (EFA) with classical latent factor models. We will show that EFA subsumes latent factor models as special cases while being able overcome their limitations in modeling complex sequential data.

We begin by reviewing latent factor models as are adopted in modeling high-dimensional discrete data~\citep{ouyang2024statistical,rudolph2016exponential}. Consider data $x := x_{1:I}$, where $x_i \in \mathbb{R}^D$. For each $i \in [I]$, let $c_i \subseteq [I] - \{i\}$ denote its \textit{context}, and we model each data point based on its context. Concretely, we model the center observation $x_i$ by $x_i \mid x_{c_i} \sim \textrm{ExpFam}(\eta_i(x_{c_i}), t(x_i))$, where $\eta_i(x_{c_i})$ and $t(x_i)$ denote the natural parameter and sufficient statistic, respectively. Latent factor models often uses the \textit{linear embedding}, defined as $$\eta_i(x_{c_i}) := f_i \left(\rho[i]^\top \sum_{j \in c_i} \alpha[j] x_j \right).$$ In this case, $f_i$ represents a predefined link function, while $\rho[i] \in \mathbb{R}^{K \times D}$ and $\alpha[i] \in \mathbb{R}^{K \times D}$ represent the center and context embeddings. These embeddings are learned via maximizing the sum of the pseudo log-likelihood across all data points, where the pseudo log-likelihood for a data point $x = x_{1:I}$ is given by $\sum_{i \in [I]} \log p(x_i \mid x_{c_i})$.

\begin{remark}[From pseudo log-likelihood to full likelihood]
    If we restrict $c_i$ to contain only indices smaller than $i$, the pseudo log-likelihood becomes the full likelihood: $$\log p(x) := \sum_{i \in [I]} \log p(x_i \mid x_{1, 2, \dots, i-1}) = \sum_{i \in [I]} \log p(x_i \mid x_{c_i}).$$
\end{remark}

In Section \ref{sec:efa-spec-case}, we show that exponential family attention (EFA) encompasses latent factor models as a special case for each specific EFA instantiation presented in Section \ref{sec:ex-of-efa}. Furthermore, in Section \ref{sec:efa-over-lim}, we compare EFA and latent factor models in detail to illustrate the added modeling capacity of EFA relative to latent factor modeling. 

\subsection{EFA includes latent factor models as a special case}
\label{sec:efa-spec-case}
In this section, we prove that EFA encompasses (linear) latent factor models as a special case. This is illustrated in Propositions \ref{prop:bert-cbow} to \ref{prop:rating}, which constructively demonstrate this relationship for EFA instantiations in Sections \ref{sec:app-baskets} to \ref{sec:app-movie}. The proofs are in Appendix \ref{app:pf-cbow} to \ref{app:pf-rating}.

\begin{proposition}[Latent factor model as a special case of EFA in Section \ref{sec:app-baskets}]
\label{prop:bert-cbow}
    Un-der a specific set of parameters, the bidirectional EFA model in Sections \ref{sec:self-attn-lang} and \ref{sec:app-baskets} reduces to a variant of linear latent factor models:
    $$x_i \mid x_{-i} \sim \textrm{Categorical}\left( \sigma \left( \frac{1}{I-1} \bm{\rho}^\top \sum_{j \in [I], j \neq i} \bm{\alpha}_{x_j} \right) \right),$$
where $\sigma$ denotes the softmax operator.
\end{proposition}

\begin{proposition}[Latent factor model as a special case of EFA in Section \ref{sec:app-gauss}]
\label{prop:gauss}
Un-der a specific set of parameters, the EFA model in Section \ref{sec:app-gauss} (with \textrm{softmax} replaced by linear attention) reduces to a variant of linear latent factor models:
$$y_i \mid y_{-i} \sim \mathcal{N}\left( (\bm{h}(\tau_i))^\top \sum_{j \in [I], j \neq i} \bm{h}(\tau_j) y_j, \sigma^2 \right).$$
\end{proposition}

\begin{proposition}[Latent factor model as a special case of EFA in Section \ref{sec:app-movie}]
\label{prop:rating}
Un-der a specific set of parameters, the simplified EFA model in Section \ref{sec:app-movie} reduces to a variant of linear latent factor models:
$$y_i - 1 \mid y_{-i} \sim \textrm{Poisson}\left( \exp\left(\frac{1}{I-1} \bm{\rho}_{x_i}^\top \sum_{j \in [I], j \neq i} \bm{\alpha}_{x_j} y_j \right)\right).$$
\end{proposition}

\subsection{EFA overcomes limitations of latent factor models}
\label{sec:efa-over-lim}
The family of latent factor models has three notable limitations: \textit{linearity}, \textit{order invariance}, and \textit{context definition}. EFA is specifically designed to address these limitations, enhancing the flexibility and expressiveness of latent factor models by incorporating the self-attention mechanism introduced by \citet{vaswani2017attention}. We elaborate on each of these limitations below.

\noindent
\textbf{Linearity.} The expression inside $f_i$ is linear, assuming an additive effect from each observation in the context. This limits the model's expressivity, making it less capable of capturing nonlinear relationships in the data. For instance, in word embeddings, the relationship between words in a sentence often hinges on their interactions rather than simply their additive contributions. Similarly, in the movie ratings example, the combined effect of ratings for two different movies by the same user on the rating of another movie may involve interactions that a simple summation cannot adequately capture.

In contrast, EFA models these interactions by considering weighted combinations of each observation's effect in the context. The weights characterize the relevance of each context observation to the center observation, with their values depending on the presence of other observation in the context. For example, in the case of word embeddings, the relevance of a word in the context to the target (center) word will vary depending on which other words surround it. This dynamic weighting allows the model to capture the complex, context-dependent relationships between words, where the contribution of each word is not fixed but instead adjusted based on the interaction with others in the context. This dependence captures the interactions among words, allowing the model to account for nonlinear effects between the context and center words.

\noindent
\textbf{Order invariance.} Latent factor models treat the order of context observations as invariant. These models describe the co-occurrence between the center and context observations by computing the inner product of the center’s latent factor and the sum (or average) of the latent factors for the context words. However, this approach inherently ignores the order in which the context words appear in the data. In other words, latent factor models assume that the ordering of words in the context has no impact on the conditional distribution of the center word. For instance, in the movie ratings example, the model assumes that the order in which a user rates movies does not affect the conditional distribution of the ratings given the context. However, this assumption is often unrealistic in practice. The order in which a user rates movies could reflect temporal or sequential preferences that are crucial for predicting future ratings.

EFA overcomes this limitation by incorporating positional encoding, which introduces a unique representation for each observations based on its position within the context. Instead of treating observations in the context as interchangeable, positional encoding ensures that the embedding of each observations depends on its specific location relative to the center observations. For instance, in natural language processing, the word ``movie" in the context of a review may carry different meanings depending on whether it appears before or after terms like ``action" or ``comedy." In the former case, ``movie" may be associated with action-related attributes, while in the latter, it may suggest comedic themes. Positional encoding ensures that these subtle shifts in meaning based on word order are captured by the model. EFA thus allows the model to distinguish between the same observations appearing in different positions within the context, capturing the varying impact that the observations have depending on their positions.

\noindent
\textbf{Context definition.} Latent factor models often require the pre-specification of the context set. For example, in word sequence modeling, we typically need to define a fixed window size around each center word in a sentence, meaning that only the $K$ words before and after the center word are considered relevant. In modeling natural language, the choice of window size is crucial, as it directly impacts model performance. A window that is too small may fail to capture essential contextual information, while a window that is too large may introduce irrelevant details. Similarly, when modeling temperature data across different U.S. cities, we must define a neighborhood for each city, which involves determining what constitutes "closeness" and selecting an appropriate number of neighbors. These pre-specification requirements can be difficult and often impractical in real-world applications.

EFA addresses this challenge by learning the context set implicitly from the data. Instead of requiring a fixed context, it treats all other observations as potential context and assigns varying weights to them based on their relevance. These weights indicate how much each context observation contributes to predicting the center observation, and they are learned through a separate latent factor model that links the context observation with the center observation. This approach allows EFA to capture more nuanced, data-driven dependencies without the need for explicit pre-specifications, making it more flexible and adaptable to a wider range of tasks.

\section{Theoretical guarantees for EFA}
\label{sec:theo-gua}

We present two theoretical results for exponential family attention (EFA). Section \ref{sec:lin-iden} proves the linear identifiability of EFA. Section \ref{sec:exc-los} establishes an excess loss generalization guarantee.

\subsection{Linear identifiability of EFA}
\label{sec:lin-iden}
This section establishes the identifiability of EFA. In Section \ref{sec:iden-efa}, we define identifiability for EFA and discuss its significance. In Section \ref{sec:iden-pf}, we prove that EFA is indeed identifiable.

\subsubsection{What identifiability means for EFA}
\label{sec:iden-efa}
Recall that in EFA, our data is of the form $(x, y) = (x_{1:I}, y_{1:I})$, where the $x_i$'s are categorical with $D$ categories and the $y_i$'s follow an exponential family distribution. EFA learns $\bm{\theta} := \bm{\theta}_1 \cup \bm{\theta}_2$ with $\bm{\theta}_1 := (\bm{\delta}, \bm{\beta}, \bm{P}, \bm{W}^\mathcal{Q}, \bm{W}^\mathcal{K}, \bm{W}^\mathcal{V})$ and
$\bm{\theta}_2 := (\bm{\bar{\delta}}, \bm{\bar{\beta}}, \bm{\bar{P}}, \bm{\bar{W}}^\mathcal{Q}, \bm{\bar{W}}^\mathcal{K}, \bm{\bar{W}}^\mathcal{V}, \bm{\lambda_1}(\cdot), \bm{\lambda_2}(\cdot))$ to maximize the following objective:
\begin{equation}
\label{eq:decomp2}
   \log p(x,y; \bm{\theta}) := \sum_{i \in [I]} \log p(x_i \mid x_{1:i-1}; \bm{\theta}_1) + \sum_{i \in [I]} \log p(y_i \mid x_{1:i}, y_{1:i-1}; \bm{\theta}_2) . 
\end{equation}
Note that the parameters $\bm{\theta}_1$ and $\bm{\theta}_2$ correspond to the first and second terms of Equation \eqref{eq:decomp2}, respectively. As $\bm{\theta}_1$ and $\bm{\theta}_2$ are disjoint, we study identifiability for each term separately.

\noindent
\textbf{First term.} Section \ref{sec:self-attn-lang} states that
$$p(x_i \mid x_{1:i-1}; \bm{\theta}_1) := \frac{\exp \left( H_{\bm{\theta}_1}(\underline{x})_i^\top e_{\bm{\theta}_1}(x_i) \right)}{\sum_{x' \in [D]}\exp \left( H_{\bm{\theta}_1}(\underline{x})_i^\top e_{\bm{\theta}_1}(x') \right)},$$
where $e_{\bm{\theta}_1}(x)$ denotes the $x$-th column of $\bm{\delta}$ and $H_{\bm{\theta}_1}(\underline{x})_i$ denotes $X_{ii}'$; we can interpret $e_{\bm{\theta}_1}(x)$ and $H_{\bm{\theta}_1}(\underline{x})_i$ as \textit{center} and \textit{context} embeddings. Intuitively, this is because $H_{\bm{\theta}_1}(\underline{x})_i$ captures all information about the context observations $x_{1:i-1}$, while $e_{\bm{\theta}_1}(x)$ captures all information about $x$ when it is a center observation. As both the center and context embeddings are represented by neural networks, which are inherently over-paramterized, it is possible to have two different $\bm{\theta}_1$'s corresponding to the same conditional distribution $p(x_i \mid x_{1:i-1}; \bm{\theta}_1)$, whence the usual notion of identifiability fails.

We thus consider the identifiability with respect to the center and context embeddings \textit{up to a linear transformation}, defined as follows:
\begin{definition}[Linear identifiability for $\bm{\theta}_1$]
\label{def:lin-id-efa-1}
    Linear identifiability of $\bm{\theta}_1$ holds if the following condition is satisfied: For any $\bm{\theta}_1', \bm{\theta}_1^* \in \Theta_1$, $p_{\bm{\theta}_1'}(x_i \mid x_{1:i-1}) = p_{\bm{\theta}_1^*}(x_i \mid x_{1:i-1})$ implies $H_{\bm{\theta}_1'}(\underline{x})_i = \bm{A}H_{\bm{\theta}_1^*}(\underline{x})_i$ and $e_{\bm{\theta}'_1}(x_i) = \bm{B}e_{\bm{\theta}^*_1}(x_i)$ for some invertible $\bm{A}, \bm{B} \in \mathbb{R}^{K \times K}$.
\end{definition}

This is the same identifiability notion as in \citet{roeder2021linear}, which we summarize in Appendix \ref{app:iden}. Definition \ref{def:lin-id-efa-1} means that for any two $\bm{\theta}_1$'s that induce the same conditional distribution, i.e., $p_{\bm{\theta}_1}(x_i \mid x_{1:i-1})$, both the center and context embeddings are identical up to linear transformations, as given by transformation matrices $\bm{A}$ and $\bm{B}$. 

\noindent
\textbf{Second term.} We focus on univariate Gaussian $y_i$'s with a known variance $\sigma^2$ (e.g., in Section \ref{sec:app-gauss}), noting that our argument can be easily adapted to the general case. From Section \ref{sec:self-attn-non-lang}, we know that
$y_i \mid x_{1:i}, y_{1:i-1}; \bm{\theta}_2 \sim \mathcal{N}(J_{\bm{\theta}_2}(\underline{y}), \sigma^2),$
where $\underline{y}$ and $J_{\bm{\theta}_2}(\underline{y})$ denote $Y_i$ and $\bm{\lambda_2}(Y_{ii}')$. We impose the following assumption on $J_{\bm{\theta}_2}(\underline{y})$.
\begin{assumption}[Structure of $J_{\bm{\theta}_2}(\underline{y})$]
\label{asm:struc}
Let $J_{\bm{\theta}_2}(\underline{y}) := K_{\bm{\theta}_2}(\underline{y})^\top L_{\bm{\theta}_2}$, where $K_{\bm{\theta}_2}(\underline{y}), L_{\bm{\theta}_2} \in \mathbb{R}^{d_j}$.
\end{assumption}

Intuitively, Assumption \ref{asm:struc} implies that $J_{\bm{\theta}_2}(\underline{y})$ can be decomposed into a product of two terms: one that depends on the other observations (i.e., the \textit{embedding} $K_{\bm{\theta}_2}(\underline{y})$) and a constant vector $L_{\bm{\theta}_2}$. Similar to Definition \ref{def:lin-id-efa-1}, we can define identifiability for $\bm{\theta}_2$ with respect to both the embedding and the constant vector.
\begin{definition}[Linear identifiability for $\bm{\theta}_2$]
\label{def:lin-id-efa-2}
    Linear identifiability of $\bm{\theta}_2$ holds if the following condition is satisfied: For any $\bm{\theta}_2', \bm{\theta}_2^* \in \Theta_2$, $p_{\bm{\theta}_2'}(y_i \mid x_{1:i}, y_{1:i-1}) = p_{\bm{\theta}_2^*}(y_i \mid x_{1:i}, y_{1:i-1})$ implies $K_{\bm{\theta}_2'}(\underline{y}) = \bm{C} K_{\bm{\theta}_2^*}(\underline{y})$ and $L_{\bm{\theta}_2'} = \bm{D}L_{\bm{\theta}_2^*}$ for some invertible $\bm{C}, \bm{D} \in \mathbb{R}^{d_j \times d_j}$.
\end{definition}

Definition \ref{def:lin-id-efa-2} means that for any two $\bm{\theta}_2$'s that induce the same conditional distribution, i.e., $p_{\bm{\theta}_2}(y_i \mid x_{1:i-i}, y_{1:i-1})$, both the embedding and the constant vector are identical up to linear transformations, as given by transformation matrices $\bm{C}$ and $\bm{D}$.

Combining Definitions \ref{def:lin-id-efa-1} and \ref{def:lin-id-efa-2}, we can define linear identifiability in the context of EFA.

\begin{definition}[Linear identifiability of EFA]
\label{def:lin-id-efa}
    Linear identifiability of EFA holds if bo-th $\bm{\theta}_1$ and $\bm{\theta}_2$ are linearly identifiable in the sense of Definitions \ref{def:lin-id-efa-1} and \ref{def:lin-id-efa-2}.
\end{definition}

\subsubsection{Proving the linear identifiability of EFA}
\label{sec:iden-pf}
Having defined linear identifiability of EFA, we now present the assumptions needed to establish that EFA is linearly identifiable.

\begin{assumption}[Diversity I]
\label{asm:div-y-efa}
    For any $\bm{\theta}_1', \bm{\theta}_1^*$ s.t. $p_{\bm{\theta}_1'}(x_i \mid x_{1:i}) = p_{\bm{\theta}_1^*}(x_i \mid x_{1:i})$, we can construct a set of $K$ distinct data tuples such that the matrices $\bm{L}'$ and $\bm{L}^*$ are invertible. Here, the columns of $\bm{L}'$ and $\bm{L}^*$ are $(e_{\bm{\theta}_1'}(x^{(j)}_{Ai}) - e_{\bm{\theta}_1'}(x^{(j)}_{Bi}))$'s and $(e_{\bm{\theta}_1^*}(x^{(j)}_{Ai}) - e_{\bm{\theta}_1^*}(x^{(j)}_{Bi}))$'s, where $j \in [K]$ and $e_{\bm{\theta}_1}(x^{(j)}_{Ai})$ (resp. $e_{\bm{\theta}_1}(x^{(j)}_{Bi})$) denotes the center embedding of the first (resp. second) element of the $j$-th data tuple.
\end{assumption}

\begin{assumption}[Diversity II]
\label{asm:div-x-efa}
    For any $\bm{\theta}_1', \bm{\theta}_1^*$ s.t. $p_{\bm{\theta}_1'}(x_i \mid x_{1:i}) = p_{\bm{\theta}_1^*}(x_i \mid x_{1:i})$, we can construct $K+1$ distinct data points s.t. $\bm{M}'$ and $\bm{M}^*$, with columns $$\left[ -\log \sum_{x' \in [D]} \exp \left( H_{\bm{\theta}_1'}(\underline{x}^{(j)})_i^\top e_{\bm{\theta}_1'}(x') \right); H_{\bm{\theta}_1'}(\underline{x}^{(j)})_i  \right]^\top$$ and $$\left[ -\log \sum_{x^* \in [D]} \exp \left( H_{\bm{\theta}_1^*}(\underline{x}^{(j)})_i^\top e_{\bm{\theta}_1^*}(x^*) \right);H_{\bm{\theta}_1^*}(\underline{x}^{(j)})_i  \right]^\top,$$ respectively, are invertible. Here, $j \in [K+1]$ and $H_{\bm{\theta}_1}(\underline{x}^{(j)})_i$ denotes the context embedding of the $j$-th data point. 
\end{assumption}

\begin{assumption}[Diversity III]
\label{asm:div-add}
    For any $\bm{\theta}_2', \bm{\theta}_2^*$ s.t. $$p_{\bm{\theta}_2'}(y_i \mid x_{1:i}, y_{1:i-1}) = p_{\bm{\theta}_2^*}(y_i \mid x_{1:i}, y_{1:i-1}),$$ we can construct $d_j$ distinct data points s.t. the matrices $\bm{N}'$ and $\bm{N}^*$ are invertible. Here, the columns of $\bm{N}'$ and $\bm{N}^*$ are $(K_{\bm{\theta}_2'}(\underline{y}^{(j)}))$'s and $(K_{\bm{\theta}_2^*}(\underline{y}^{(j)}))$'s, where $j \in [d_j]$ and $K_{\bm{\theta}_2}(\underline{y}^{(j)})$ denotes the embedding of the $j$-th data point. 
\end{assumption}

Intuitively, Assumptions \ref{asm:div-y-efa} to \ref{asm:div-add} are imposed to guarantee the existence of diverse sets of data points, which is crucial for ensuring the identifiability of the model parameters. Similar diversity assumptions are commonly imposed in the literature \citep{roeder2021linear}.

\begin{remark}[On the satisfiability of Assumptions \ref{asm:div-y-efa} to \ref{asm:div-add}]
    We need $K \leq D - 1$ for Assumption \ref{asm:div-y-efa} to hold (see Section 3.3 of \citet{roeder2021linear}). Otherwise, Assumptions \ref{asm:div-y-efa} to \ref{asm:div-add} are mild in the context of deep neural networks and should simultaneously hold almost surely.
\end{remark}

We now present our main result, whose proof is in Appendix \ref{app:pf-iden}.

\begin{theorem}[Linear identifiability of EFA]
\label{thm:iden}
    Under Assumptions \ref{asm:struc} to \ref{asm:div-add}, EFA is linearly identifiable in the sense of Definition \ref{def:lin-id-efa}.
\end{theorem}
To emphasize, linear identifiability here refers to the uniqueness of the center and context embeddings for $\bm{\theta}_1$, as well as the embedding for $\bm{\theta}_2$, up to linear transformations.

\subsection{Excess loss generalization guarante for EFA}
\label{sec:exc-los}

In this section, we establish an excess loss generalization guarantee for exponential family attention (EFA). We focus on a simplified version of the Gaussian spatiotemporal time series example in Section \ref{sec:app-gauss}, although the results can be readily extended to other EFA applications. We consider the same transformer model as in \citet{bai2024transformers} for theroetical convenience. We first outline the model and loss function in Section \ref{sec:arch}. We then introduce the theoretical setup and key assumptions in Section \ref{sec:theo-erm} and present the main result in Section \ref{sec:theo-result}.

\subsubsection{Model, parameters, and loss function}
\label{sec:arch}
\noindent
\textbf{Model.} The training data is represented as $Z := \{(\tau_1, y_1), \cdots, (\tau_I, y_I)\}$, where $\tau_i \in \mathbb{R}^\tau$ and $y_i \in \mathbb{R}$ for each $i \in [I]$. For simplicity, we omit the time index $f$ since each training data consists of observations from the same time index. We assume that $\bm{g}(\tau_i) := \mathbf{G}_2 \sigma(\mathbf{G}_1 \tau_i) \in \mathbb{R}^K$ for each $i \in [I]$, where $\mathbf{G}_1 \in \mathbb{R}^{K' \times \tau}$ and $\mathbf{G}_2 \in \mathbb{R}^{K \times K'}$ are learned and $\sigma$ denotes the ReLU activation function. For each observation $Z$, we aim to model $y_i$ conditional on $y_{-i}$ under the following model.
\begin{enumerate}[leftmargin=0.6cm]
\item Let $$
Y_i := 
\left[
  \begin{array}{ccccccc}
    \bm{g}(\tau_{1}) & \ldots & \bm{g}(\tau_{i-1}) & \bm{g}(\tau_{i}) & \bm{g}(\tau_{i+1}) & \ldots & \bm{g}(\tau_{I}) \\
    y_1 & \ldots & y_{i-1} & 0 & y_{i+1} & \ldots & y_{I}
  \end{array}
\right] \in \mathbb{R}^{(K+1) \times I}.
$$
This model represents a special case of EFA where $\bm{\lambda_1}(y) = y$ when $y \neq \texttt{[MASK]}$ and $\bm{\lambda_1}(\texttt{[MASK]}) = 0$. Also, as we wish to predict $y_i$, we replace it with the the \texttt{[MASK]} token.
\item Apply an $M$-head, $L$-layer transformer model and obtain $Y_i' \in \mathbb{R}^{(K+1) \times I}$.
\item The masked observation $y_i$ satisfies
$$y_i \mid y_{-i} \sim \mathcal{N}\left((Y'_{ii})_{K+1}, \sigma^2 \right)$$
for some variance $\sigma^2$. Here, $(Y'_{ii})_{K+1}$ represents the $(K+1,i)$-th entry of $Y_i'$. Note that this is a special case of EFA where $\bm{\lambda_2}(\cdot)$ is equivalent to reading off the last entry of a vector.
\end{enumerate}

\noindent
\textbf{Parameters.} The parameters in this model are $\bm{\theta} = (\bm{\theta}^{1:L}_{\texttt{attn}}, \bm{\theta}^{1:L}_{\texttt{mlp}}, \mathbf{G}_1, \mathbf{G}_2)$, where $\bm{\theta}^{1:L}_{\texttt{attn}}$ and $\bm{\theta}^{1:L}_{\texttt{mlp}}$ are the attention and MLP model parameters, and $\mathbf{G}_1$ and $\mathbf{G}_2$ are the parameters corresponding to the function $\bm{g}$.

\noindent
\textbf{Loss formulation.} Suppose we draw $F$ training instances $Z^f := \{(\tau^f_1, y^f_1), \cdots, (\tau^f_I, x^f_I)\}$ by first sampling a data distribution $P \sim \pi$, followed by drawing examples $\{(\tau^f_i, y^f_i)\}_{i=1}^I$ from $P$. For instance, in the zebrafish neurons example studied by \citet{rudolph2016exponential}, $\pi$ can represent the distribution of neuron activities at different time points. During training, the goal is to minimize
$\hat{L}(\bm{\theta}) := \frac{1}{F} \sum_{f=1}^F \ell(\bm{\theta}; Z^f),$
where
$$\mathbb{\ell}(\bm{\theta}; Z^f) := \frac{1}{2I} \sum_{i=1}^I \left( y^f_{i} - \left(Y^{f'}_{ii}\right)_{K+1} \right)^2.\footnote{Note that maximizing the Gaussian log-likelihood is equivalent to minimizing the mean squared error.}$$

\subsubsection{Theoretical setup and assumptions}
\label{sec:theo-erm}
We begin with imposing a boundedness assumption on the $y_i$'s.
\begin{assumption}[Boundedness of the $y_i$'s]
\label{asm:asm-1}
    We have $|y_i| \leq B_y$ for each $i \in [I]$. 
\end{assumption}
This is a common assumption imposed for theoretical convenience \citep{bai2024transformers}. In addition, we slightly modify the loss function by introducing the $\textrm{clip}$ operator to bound the magnitude of each value within a pre-defined range. Concretely, we have
$$\mathbb{\ell}(\bm{\theta}; Z^f) := \frac{1}{2I} \sum_{i=1}^I \left( y_i^f - \textrm{clip}_{B_y}(\textrm{read}_y(\textrm{TF}_{\bm{\theta}}^R(\textrm{Exp}_{\bm{\theta}}^R(Z^f_{i})))) \right)^2,$$
where $\textrm{read}_y$ and $\textrm{TF}_{\bm{\theta}}^R$ respectively denote the output of reading off the last entry of the masked column and repeatedly applying the clipped attention and MLP layers, and $$\textrm{Exp}_{\bm{\theta}}^R(Z_i^f) := \left[
  \begin{array}{ccccccc}
    \bm{\tilde{g}}(\tau^f_{1}) & \ldots & \bm{\tilde{g}}(\tau^f_{i-1}) & \bm{\tilde{g}}(\tau^f_{i}) & \bm{\tilde{g}}(\tau^f_{i+1}) & \ldots & \bm{\tilde{g}}(\tau^f_{I}) \\
    y_1 & \ldots & y_{i-1} & 0 & y_{i+1} & \ldots & y_I
  \end{array}
\right] \in \mathbb{R}^{(K+1) \times N},$$
where $\bm{\tilde{g}}(\tau^f_{i}) =  \mathbf{G}_2 \sigma(\mathbf{G}_1 \cdot \textrm{clip}_R(\tau^f_{i})) \in \mathbb{R}^K$ for each $i \in [I]$. In practice, $R$ can be chosen to be as large as possible so that the behavior of the transformer is not altered by the $\textrm{clip}$ operator.

\noindent
\textbf{Empirical risk minimization.} We frame the learning problem for EFA as a standard empirical risk minimization (ERM) problem. First, denote the population quantity as
$$L(\bm{\theta}) = \mathbb{E}_{P \sim \pi, Z \sim P}\left[\ell(\bm{\theta}; Z)\right].$$
The learning problem is as follows:
% \begin{align*}
    $$\hat{\bm{\theta}} := \argmin_{\bm{\theta} \in \Theta_{L, M, D', K', B}} \hat{L}(\bm{\theta}),$$ where 
    $$\Theta_{L, M, D', K', B} := \left\{ \bm{\theta}: \max_{\ell \in [L]} M^{(\ell)} \leq M, \max_{\ell \in [L]} D^{(\ell)} \leq D', K^{(0)} \leq K', \vertiii{\bm{\theta}} \leq B\right\}.$$ Here,
% \end{align*}
\begin{align*}
    \vertiii{\bm{\theta}} &:= \max_{\ell \in [L]} \left\{ \max_{m \in [M]} \left\{ \| \mathbf{Q}_m^{(\ell)} \|_{\text{op}}, \| \mathbf{K}_m^{(\ell)} \|_{\text{op}} \right\} + \sum_{m=1}^M  \| \mathbf{V}_m^{(\ell)} \|_{\text{op}} + \| \mathbf{W}_1^{(\ell)} \|_{\text{op}} + \| \mathbf{W}_2^{(\ell)} \|_{\text{op}}  \right\} \\
&+ \| \mathbf{G}_1 \|_{\text{op}} + \| \mathbf{G}_2 \|_{\text{op}},
\end{align*}
where $\| \cdot\|_{\textrm{op}}$ denotes the operator norm, $M^{(\ell)}$ denotes the number of heads on the $\ell$-th attention layer, $D^{(\ell)}$ denotes the first dimension of $\mathbf{W}_1^{(\ell)}$ (i.e., the MLP layer), and $K^{(0)}$ denotes the first dimension of $\mathbf{G}_1$. 

\subsubsection{Theoretical result}
\label{sec:theo-result}
We next establish an excess loss generalization guarantee for EFA. The proof is in Appendix \ref{pf:theo}.
\begin{theorem}[Generalization guarantee for EFA]
\label{thm:gen-gua}
Suppose Assumption \ref{asm:asm-1} is satisfied. Then, with a probability of at least $1 - \xi$ over the training data $\{\bm{Z}^f\}_{f \in [F]}$, the solution $\hat{\bm{\theta}}$ to the ERM problem in Section \ref{sec:theo-erm} satisfies
$$L(\hat{\bm{\theta}}) \leq \inf_{\bm{\theta} \in \Theta_{L, M, D', K', B}} L(\bm{\theta}) + O\left(B_y^2 \sqrt{\frac{43L(L(3MD^2 + 2DD') + K'(K + \tau)) \iota + \log \left(\frac{1}{\xi}\right) }{F}} \right),$$
where $\iota := \log \left( 2 + \max \left( B, R, \frac{1}{2B_y} \right)\right)$
\end{theorem}

Theorem \ref{thm:gen-gua} provides an upper bound on the difference between $L(\hat{\bm{\theta}})$ (i.e., the population loss wrt. the learned $\hat{\bm{\theta}}$) and $\inf_{\bm{\theta} \in \Theta_{L, M, D', K', B}} L(\bm{\theta})$ (i.e., the minimum population loss wrt. all possible $\bm{\theta}$ within $\Theta_{L, M, D', K', B}$). Intuitively, as the sample size $F$ increases, the learned embedding function $\hat{\bm{g}}(\tau) = \mathbf{\hat{G}_2} \sigma(\mathbf{\hat{G}_1} \tau)$ becomes more accurate in the sense that the corresponding population loss approaches its theoretical minimum. Moreover, if the actual data generation process closely follows the structure of the EFA model, $\inf_{\bm{\theta} \in \Theta} L(\bm{\theta})$ will be small, which implies that $L(\hat{\bm{\theta}})$ will also be small. This result is akin to Theorem K.1 of \citet{bai2024transformers} but extends to EFA.

% !TeX root = ./main.tex
\section{Experiments}
\label{sec:real-data}
In this section, we demonstrate the effectiveness of exponential family attention (EFA), compared with linear latent factor models (FM) outlined in Propositions \ref{prop:bert-cbow}, \ref{prop:gauss}, and \ref{prop:rating}. This is shown via experiments on a synthetic data set (Section \ref{sec:synth}) and three real-world data sets: MovieLens ratings (Section \ref{sec:movie}), Instacart baskets (Section \ref{sec:groc}), and U.S. cities' temperatures (Section \ref{sec:temp}). Experimental details and model instantiations are in Appendix \ref{app:ex}.

\subsection{Synthetic data}
\label{sec:synth}
To highlight the advantages of EFA over FM, we conduct a synthetic data experiment where we model movie ratings based on the ratings of previously rated movies (unidirectional) or all other rated movies (bidirectional). We design a data generation process that includes position-dependent signals to illustrate the modeling power of EFA; see \Cref{app:synth-details} for details.

\noindent
\textbf{Models.} The FM baseline follows the model structure in Proposition \ref{prop:rating}, with the $\exp(\cdot)$ layers removed and the Poisson distribution replaced with the Gaussian distribution with a known variance\footnote{This is the bidirectional version. The unidirectional version can be obtained by replacing $-i$ with $<i$ and $j \neq i$ with $j < i$.}:
$y_i  \mid y_{-i} \sim \mathcal{N}\left( \frac{1}{I-1} \bm{\rho}_{x_i}^\top \sum_{j \in [I], j \neq i} \bm{\alpha}_{x_j} y_j, \sigma^2\right).$
Moreover, we use 32-dimensional embeddings. For EFA, we learn 32-dimensional categorical embeddings for both movies ($\bm{\bar{\beta}}$) and ratings ($\bm{\lambda_1}$), along with a special embedding for the masked rating. These embeddings are concatenated and passed through a 2-layer, 2-head self-attention mechanism. Finally, a series of non-linear dense layers ($\bm{\lambda_2}$) are applied.

\begin{table}[t]
\centering
\caption{EFA outperforms FM in both unidirectional and bidirectional settings, using synthetic data for predicting movie ratings (\textit{left}, Section \ref{sec:synth}) and MovieLens data for predicting sequences of movies (\textit{right}, Section \ref{sec:exp-1}).}
\begin{minipage}{0.4\textwidth}
    \centering
    \begin{tabular}{ccc}
    \hline
    Setting & Model & Test MSE \\
    \hline
    \multirow{2}{*}{Unidirectional} & FM & 4.519 \\
    \cline{2-3}
    & EFA & \textbf{1.033} \\
    \hline
    \multirow{2}{*}{Bidirectional} & FM & 2.636 \\
    \cline{2-3}
    & EFA & \textbf{1.038} \\
    \hline
\end{tabular}
\end{minipage}
\hspace{0.03\textwidth} % Adjust the gap size here
\begin{minipage}{0.55\textwidth}
    \centering
    \begin{tabular}{ccc}
    \hline
    Setting & Model & Test cross-entropy \\
    \hline
    \multirow{2}{*}{Unidirectional} & FM & 3.534 \\
    \cline{2-3}
    & EFA & \textbf{3.444} \\
    \hline
    \multirow{2}{*}{Bidirectional} & FM & 3.566 \\
    \cline{2-3}
    & EFA & \textbf{3.483} \\
    \hline
\end{tabular}
\end{minipage}
\label{tab:}
\end{table}

\noindent
\textbf{Results.} In this setting, maximizing the (pseudo) log-likelihood is equivalent to minimizing the mean squared error (MSE). Table \ref{tab:} shows that EFA yields significantly lower MSEs as compared to FM in both unidirectional and bidirectional scenarios.

\subsection{MovieLens ratings}
\label{sec:movie}

We conduct two separate experiments on the MovieLens data due to a decomposition of the EFA log-likelihood. Specifically, the EFA log-likelihood (Equation \eqref{eq:ll}) decomposes into two disjoint optimization problems: one for $\sum_{i \in [I]} \log p(x_i \mid x_{1:i-1}; \bm{\theta})$, which models the sequence of movies rated by each user, and another for $\sum_{i \in [I]} \log p(y_i \mid x_{1:i}, y_{1:i-1}; \bm{\theta})$, which models the ratings assigned to these movies given the set of rated movies. This decomposition allows us to conduct two separate experiments: (a) predicting the sequence of movies rated by each user (Section \ref{sec:exp-1}), and (b) predicting the ratings assigned to movies, given the set of rated movies (Section \ref{sec:exp-2}). For both tasks, we found that EFA consistently outperforms FM.

\subsubsection{Experiment 1: Modeling sequences of movies}
\label{sec:exp-1}
\noindent
\textbf{Setup.} We use the MovieLens 100K data available through GroupLens,\footnote{\url{https://grouplens.org/datasets/movielens/100k/}} which contains 100,000 ratings from 1,000 users on 1,700 movies. Each rating is accompanied by a timestamp indicating when the user rated the movie. However, many users tend to rate multiple movies at the same time, resulting in a significant number of duplicate timestamps. To address this, we first select the top 50 movies based on the number of unique users who rated them. Next, we remove users whose number of reviews is at least twice the number of unique timestamps. This step ensures that we exclude users who tend to rate multiple movies simultaneously, preserving the temporal structure of the data. For each remaining user, we randomly select one movie corresponding to each timestamp. The final data set consists of 902 users and 50 movies, with a sparsity of 69.50\%.

\noindent
\textbf{Models.} For the FM baseline, we use the model described in Proposition \ref{prop:bert-cbow}, where each movie is represented by 32-dimensional embeddings. We evaluate two setups: using all other movies as context (bidirectional) and using only previously rated movies as context (unidirectional). For comparison, we also employ an EFA variant with a 2-head, 2-layer self-attention mechanism and 32-dimensional embeddings, using the same setups as in FM. We randomly split the users into training, validation, and test sets in proportions of 56.25\%, 18.75\%, and 25\%, respectively.

\noindent
\textbf{Optimization.} We employ the Adam optimizer \citep{kingma2014adam} with a learning rate of $10^{-4}$ for up to 2,000 epochs and apply early stopping with a patience of 10 epochs.

\noindent
\textbf{Results.} Table \ref{tab:} shows that the test categorical cross-entropy losses are lower for EFA as compared to FM in both unidirectional and bidirectional cases. This illustrates the improvement of EFA over FM in modeling the sequences of rated movies.

\subsubsection{Experiment 2: Modeling ratings assigned by each user}
\label{sec:exp-2}
\noindent
\textbf{Setup.} We retain only ratings of 3, 4, or 5, subtracting 2 from each before applying the same preprocessing steps as in the first experiment, following \citet{rudolph2016exponential}. The resulting data set includes 893 users and 50 movies, with a sparsity of 72.08\%.

\begin{table}[t]
\caption{EFA achieves better performance than FM across all four settings. Here, the mean predicted ratings correspond to actual ratings of 1, 2 and 3, respectively.}
\vspace{5mm}
\centering
\begin{tabular}{cccccc}
\hline
\multicolumn{2}{c}{\multirow{2}{*}{Setting}} & \multicolumn{2}{c}{Test cross-entropy} & \multicolumn{2}{c}{Mean predicted ratings} \\ \cline{3-6} 
\multicolumn{2}{c}{}                         & FM                  & EFA                  & FM                  & EFA                 \\ \hline
\multirow{2}{*}{v1}     & Bidirectional      & 0.971                 & \textbf{0.945}                 & (2.06, 2.10, 2.21)   & (1.93, 2.06, 2.30)  \\ \cline{2-6} 
                        & Unidirectional     & 0.972                 & \textbf{0.953}                 & (2.06, 2.09, 2.20)   & (1.98, 2.09, 2.29)  \\ \hline
\multirow{2}{*}{v2}     & Bidirectional      & 0.507                 & \textbf{0.494}                 & (2.06, 2.10, 2.21)   & (1.94, 2.08, 2.31)  \\ \cline{2-6} 
                        & Unidirectional     & 0.508                 & \textbf{0.499}                 & (2.06, 2.10, 2.20)   & (1.98, 2.09, 2.29)  \\ \hline
\end{tabular}
\label{tab:exp3}
\end{table}

\noindent
\textbf{Models.} The FM baseline follows the model structure in Proposition \ref{prop:rating}, with two different parameterizations (named as v1 and v2)\footnote{This is the bidirectional version. The unidirectional version can be obtained by replacing $-i$ with $<i$ and $j \neq i$ with $j < i$.}:
$$y_i - 1 \mid y_{-i} \sim \textrm{Poisson}\left( \exp\left(\frac{1}{I-1} \bm{\rho}_{x_i}^\top \sum_{j \in [I], j \neq i} \bm{\alpha}_{x_j} y_j \right)\right);$$
$$y_i  \mid y_{-i} \sim \textrm{Poisson}\left( 1 + \exp\left(\frac{1}{I-1} \bm{\rho}_{x_i}^\top \sum_{j \in [I], j \neq i} \bm{\alpha}_{x_j} y_j \right)\right).$$ Moreover, the FM uses 32-dimensional embeddings. For EFA, we learn 32-dimensional categorical embeddings for both movies ($\bm{\bar{\beta}}$) and ratings ($\bm{\lambda_1}$), along with a special embedding for the masked rating. These embeddings are concatenated and passed through a 2-layer, 2-head self-attention mechanism. Finally, a series of non-linear dense layers ($\exp(\bm{\lambda_2})$) are applied.

\noindent
\textbf{Optimization.} We maximize the (pseudo) log-likelihood utlizing the Adam optimizer \citep{kingma2014adam} with a learning rate of $10^{-4}$ for up to 1,000 epochs. We apply early stopping on the validation set, with a patience of 10 epochs.

\noindent
\textbf{Results.} We evaluate four settings: \{v1, v2\} $\times$ \{bidirectional, unidirectional\}. Table \ref{tab:exp3} summarizes the test set performance of FM and EFA across all four settings. EFA consistently achieves lower (pseudo) negative log-likelihood (i.e., cross-entropy) values compared to FM, indicating better predictive performance. Additionally, we compare the mean predicted ratings for actual ratings of 1, 2, and 3, respectively. In all settings, the predicted values for EFA are closer to (1, 2, 3) than those for FM, further demonstrating the superiority of EFA.

\subsection{Instacart baskets}
\label{sec:groc}
\noindent
\textbf{Setup.} We use Instacart's market basket data available from Kaggle.\footnote{\url{https://www.kaggle.com/c/instacart-market-basket-analysis/data}} The data set contains approximately 3.2 million baskets in the training set and 130,000 baskets in the test set. For our analysis, we focus on 63 items that appear in at least 50,000 different baskets. Additionally, we only include baskets containing at least 10 of these items. The final data set consists of 24,781 training baskets and 1,433 test baskets.
\begin{table}[t]
\caption{Top three items frequently purchased together with four representative items obtained using each method.}
\vspace{5mm}
\scriptsize
\begin{tabular}{|p{1.3cm}|p{3.15cm}|p{3.15cm}|p{3.15cm}|p{3.15cm}|}
\hline
Item   & Bidirectional FM  & Unidirectional FM   & Bidirectional EFA  & Unidirectional EFA\\ \hline
Asparagus         & Fresh Cauliflower, Chicken Breasts, Organic Zucchini & Fresh Cauliflower, Raspberries, Organic Zucchini   & Yellow Onions, Organic Yellow Onions, Organic Zucchini                 & Organic Garlic, Half \& Half, Organic Baby Carrots            \\ \hline
Carrots           & Organic Celery, Organic Parsley, Yellow Onions       & Organic Celery, Yellow Onions, Original Hummus     & Organic Avocado, Organic Blackberries, Half \& Half                    & Organic Baby Carrots, Organic Garlic, Organic Red Onions      \\ \hline
Fresh Cauliflower & Organic Zucchini, Asparagus, Organic Raspberries     & Organic Zucchini, Asparagus, Organic Yellow Onions & Organic Baby Spinach, Extra Virgin Olive Oil, Organic Red Black Pepper & Organic Zucchini, Organic Yellow Onions, Organic Raspberries  \\ \hline
Organic Avocado   & Banana, Organic Baby Spinach, Cucumber Kirby         & Organic Cilantro, Blueberries, Limes               & Organic Cilantro, Organic Garlic, Red Vine Tomato                      & Organic Bananas, Organic Baby Spinach, Extra Virgin Olive Oil \\ \hline
\end{tabular}
\label{tab:comp}
\end{table}
\noindent
\textbf{Models.} For the FM baseline, we use the model described in Proposition \ref{prop:bert-cbow}, where each item is represented by 32-dimensional embeddings. We evaluate two setups: using all other items in the same basket as context (bidirectional) and using only items previously added to the basket as context (unidirectional). For comparison, we also employ an EFA variant with a 2-head, 2-layer self-attention mechanism and 32-dimensional embeddings, using the same setups as in FM.

\noindent
\textbf{Optimization.} In all experiments, we minimize the categorical cross-entropy loss using the Adam optimizer \citep{kingma2014adam} with a learning rate of $10^{-4}$ for 1,000 epochs. The validation set comprises 25\% of the training baskets. For all instances, the cross-entropy losses on the validation sets indicate that the models have successfully converged.

\noindent
\textbf{Results.} The test categorical cross-entropy losses demonstrate that EFA outperforms FM, both when using all other items in the same basket as context (3.420 vs. 3.476) and when considering only the items previously added to the basket as context (3.706 vs. 3.715).

\begin{figure}[t]
\centering
% First row with two images
\begin{minipage}{0.49\textwidth}
    \centering
    \includegraphics[width=\textwidth]{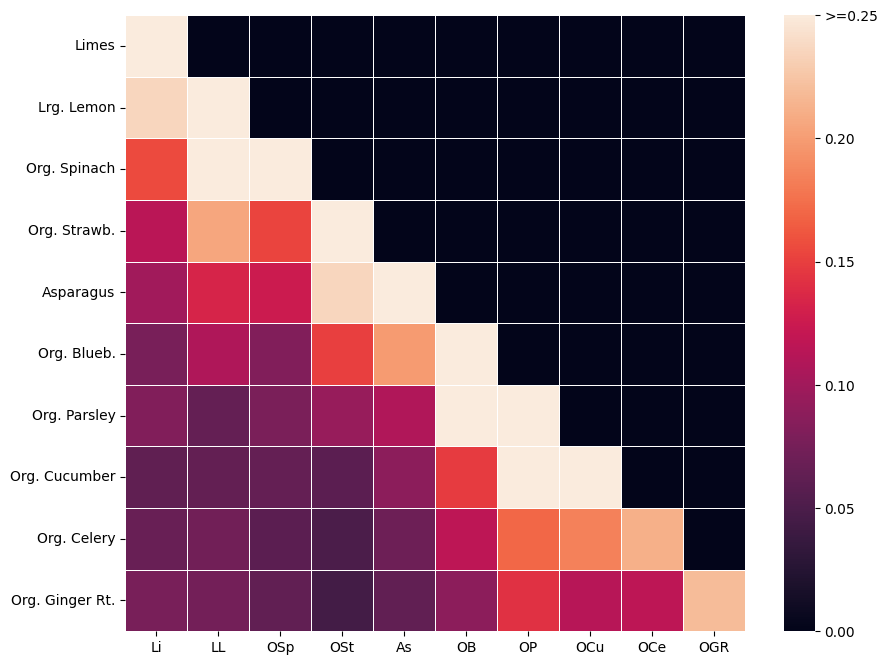}
    \subcaption{\nth{1} basket, \nth{1} layer}  
    \label{fig:layer1-item1}
\end{minipage}%
\begin{minipage}{0.49\textwidth}
    \centering
    \includegraphics[width=\textwidth]{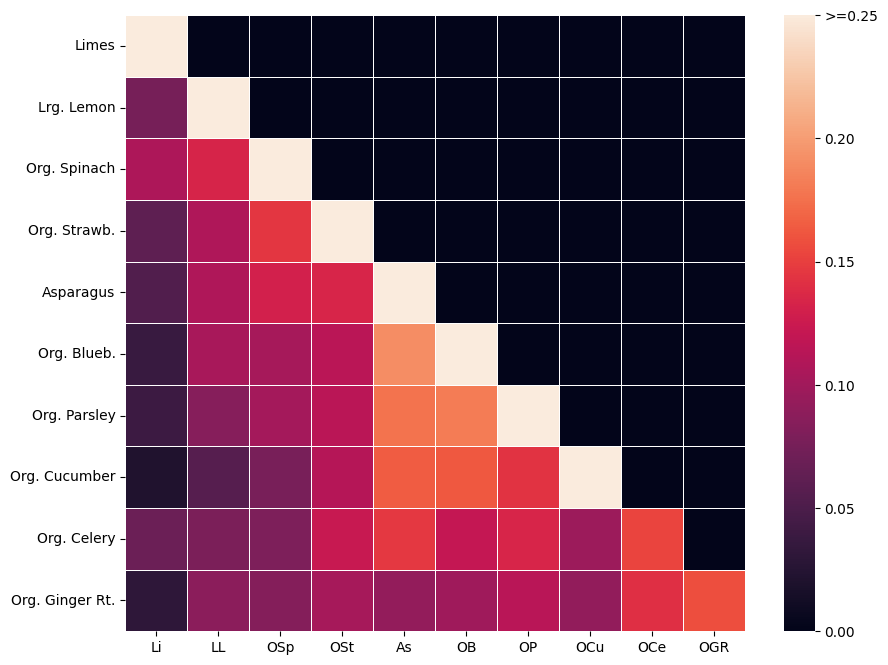}
    \subcaption{\nth{1} basket, \nth{2} layer} 
    \label{fig:layer2-item1}
\end{minipage}

% Second row with two images for item2
\begin{minipage}{0.49\textwidth}
    \centering
    \includegraphics[width=\textwidth]{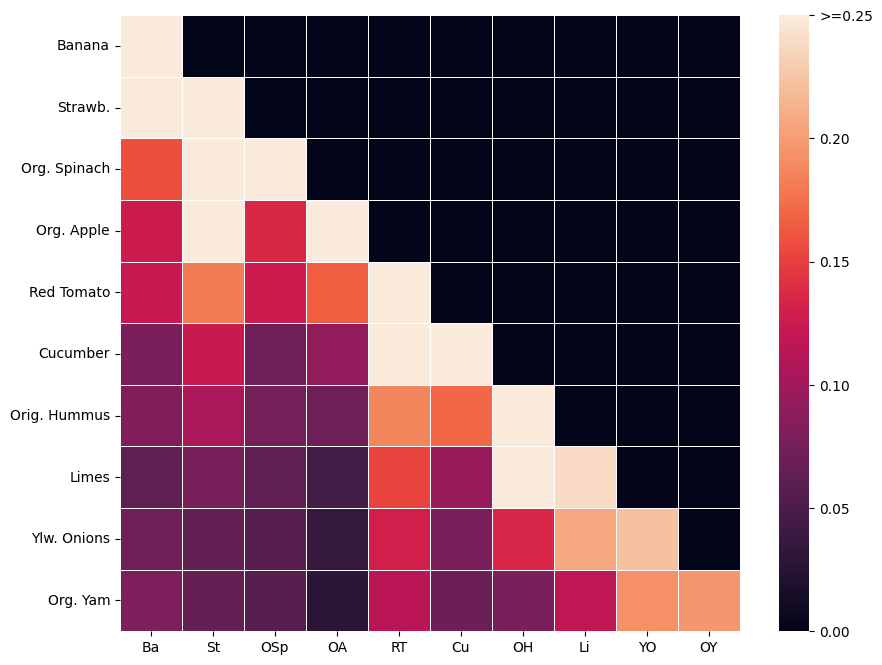}
    \subcaption{\nth{2} basket, \nth{1} layer}  
    \label{fig:layer1-item2}
\end{minipage}%
\begin{minipage}{0.49\textwidth}
    \centering
    \includegraphics[width=\textwidth]{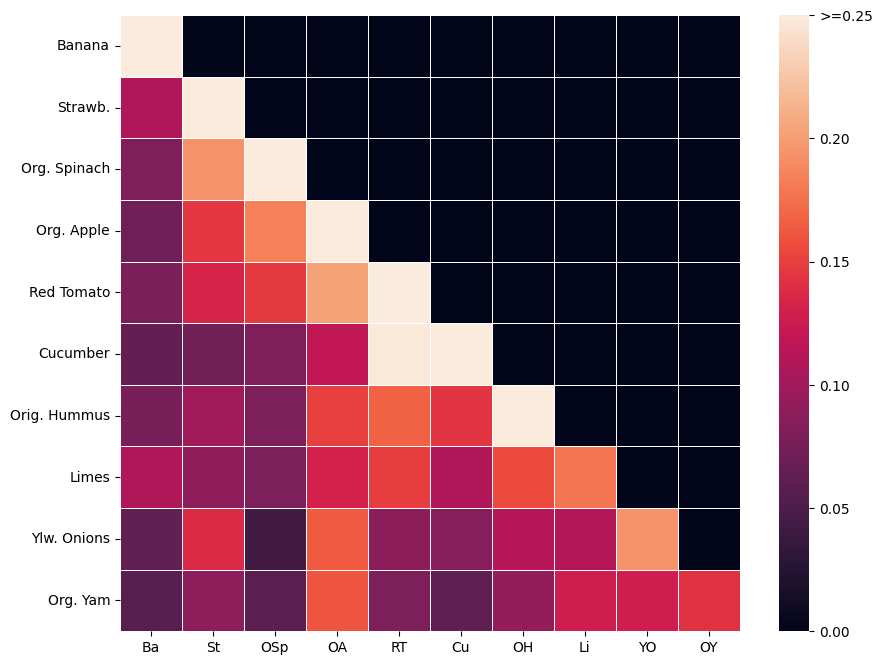}
    \subcaption{\nth{2} basket, \nth{2} layer}  
    \label{fig:layer2-item2}
\end{minipage}

\caption{Unidirectional attention weights for two representative baskets.}
\label{fig:viz-weight}
\end{figure}

Moreover, Table \ref{tab:comp} summarizes the top three items frequently purchased together with four representative items obtained using each method. Specifically, the top three items frequently purchased together with item $\Delta$ are defined as the three items $\Gamma \neq \Delta$ that maximize $\delta_{\Delta}^\top \beta_{\Gamma} + \beta_{\Gamma}^\top \delta_{\Delta}$ \citep{chen2019joint}, where $\delta_{\bm{\star}}$ and $\beta_{\bm{\star}}$ respectively denote the center and context embeddings of item $\bm{\star}$. We observe that EFA not only achieves a better fit to the data---evidenced by its lower cross-entropy loss---but also identifies items frequently purchased together that differ from those suggested by FM. This difference potentially reflects EFA's enhanced capacity to capture nuanced relationships between items.

To illustrate the enhanced modeling capabilities of EFA, we present two visualizations. Figure \ref{fig:viz-weight} shows the unidirectional attention weights for each layer, i.e.,
$$\textrm{softmax}\left(\frac{\left( \bm{W}^\mathcal{Q} X_i \right)^\top \bm{W}^\mathcal{K} X_i}{K} + M \right),$$ averaged across all attention heads for two representative baskets. In contrast to FM, which assigns uniform weights to all previous items in a basket (i.e., each row of the heat map having the same color), EFA dynamically computes the attention weights for each item based on all previous items in the same basket, resulting in varied attention patterns across layers and baskets. This enables EFA to capture complex dependencies between items in each basket. 

\begin{figure}[t]
\centering
\begin{minipage}{0.49\textwidth}
    \centering
    \includegraphics[width=\textwidth]{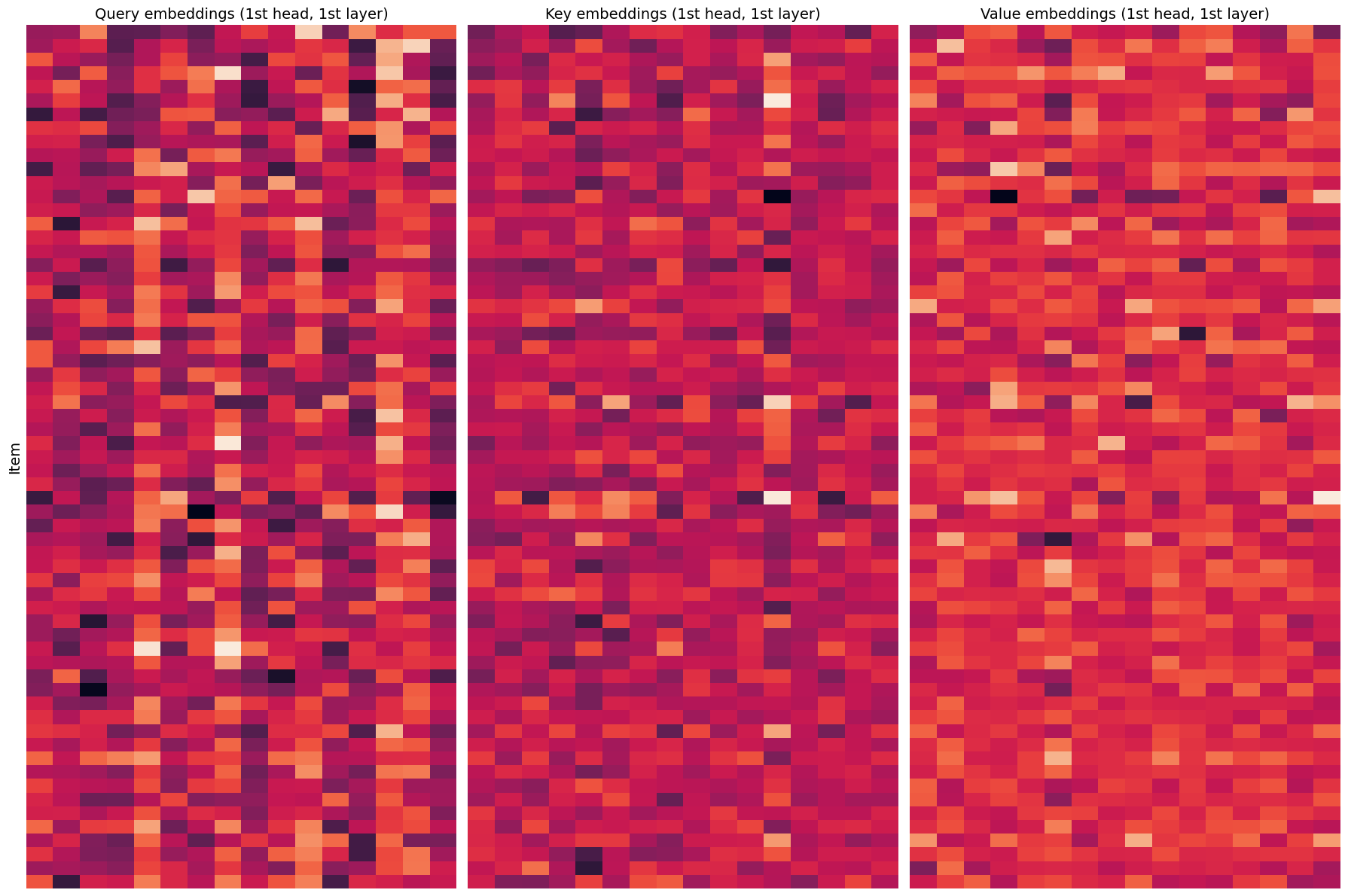}
    \subcaption{$Q$, $K$, $V$ embeddings (\nth{1} head, \nth{1} layer)}
    
\end{minipage}%
\begin{minipage}{0.49\textwidth}
    \centering
    \includegraphics[width=\textwidth]{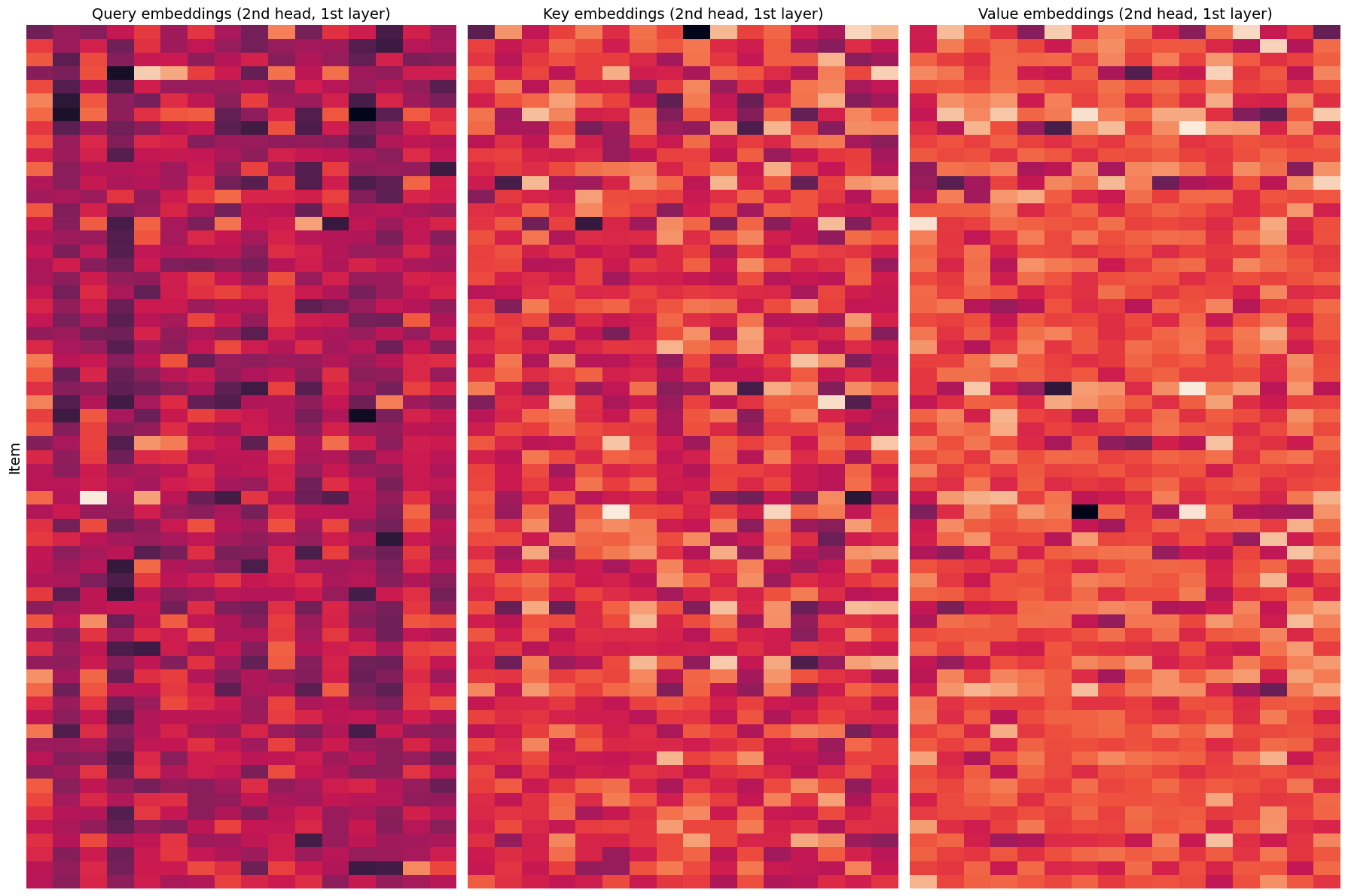}
    \subcaption{$Q$, $K$, $V$ embeddings (\nth{2} head, \nth{1} layer)}
\end{minipage}%
\caption{Per-item query, key, and value embeddings for each head of the first layer.}
\label{fig:viz-emb}
\end{figure}

Finally, Figure \ref{fig:viz-emb} presents the learned query, key, and value (QKV) embeddings (i.e., $\bm{W}^\mathcal{Q} \bm{\beta}$, $\bm{W}^\mathcal{K} \bm{\beta}$, and $\bm{W}^\mathcal{V} \bm{\beta}$) for each item, corresponding to each head of the first layer. In contrast to FM, which only uses context embeddings $\bm{\beta}$, EFA's learned QKV embeddings facilitate more nuanced interactions between the query, key, and value vectors, yielding refined context representations. This enables the model to capture richer, higher-order relationships between items.

\subsection{U.S. cities temperatures}
\label{sec:temp}
\begin{figure}[t]
\centering
\includegraphics[width=14cm]{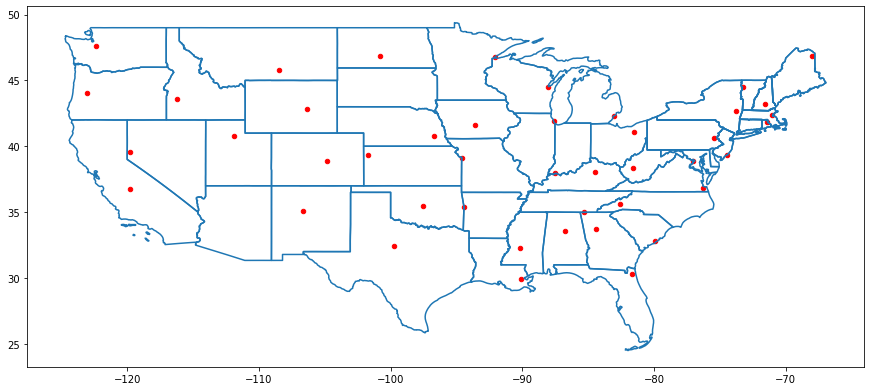}
\caption{All cities included in the experiment in Section \ref{sec:temp}, as indicated by dots.}
\label{fig:map}
\end{figure}

\noindent
\textbf{Setup.} We use the temperature data made available by the University of Dayton.\footnote{\url{https://www.kaggle.com/datasets/sudalairajkumar/daily-temperature-of-major-cities}} We focus on daily temperature records in October from 2007 to 2019. To preprocess the data, we begin by selecting one city from each state in the contiguous United States, excluding Arizona, Connecticut, South Dakota, and Delaware due to the presence of missing data. The resulting data set is then divided into three subsets: training (2007-2012), validation (2013-2016), and test (2017-2019) sets. Figure \ref{fig:map} illustrates the 44 cities (in 44 states) we consider in our analysis. 

\noindent
\textbf{Models.} For FM, we adopt the model structure outlined in Proposition \ref{prop:gauss}, except that the summation is taken over the $k \in \{1, 2, 3, 4, 5, 10, 15, 20, 25, 30, 35, 40\}$ closest time series from $i$, where ``closeness" is determined via the Haversine distance between pairs of cities. Although the time steps are not uniformly spaced---as we only consider temperatures in October---the model remains applicable because the context for each observation only consists of observations from the same time step. In addition, $\bm{h}$ is a non-linear mapping that transforms each city's coordinates (latitude and longitude) into a 32-dimensional embedding.

For EFA, we do not set a fixed number of neighbors and instead allow the temperature of a city to depend on the temperatures of all 43 other cities. We learn two non-linear mappings: one that transforms each city's coordinates (latitude and longitude) into a 32-dimensional embedding ($\bm{g}$), and another that maps each temperature to a 32-dimensional embedding ($\bm{\lambda_1}$). In addition, we learn a special embedding for the masked temperature. These two embeddings are concatenated before applying an $\ell$-layer, 2-head self-attention mechanism, where $\ell \in \{1, 2, 4\}$. Finally, a series of non-linear dense layers ($\bm{\lambda_2}$) are applied to the masked temperature's token.

\noindent
\textbf{Optimization.} Since $y_i \mid y_{-i} \sim \mathcal{N}(\bm{\lambda_2}(Y'_{ii}), \sigma^2)$ for both models, the log likelihood for one observation is
$-\frac{1}{2} \log(2\pi \sigma^2) - \frac{\left(y_i - \bm{\lambda_2}(Y'_{ii}) \right)^2}{2\sigma^2}.$
Thus, maximizing the pseudo log-likelihood is equivalent to minimizing the mean squared error (MSE). To minimize the MSE, we utilize the Adam optimizer \citep{kingma2014adam} with a learning rate of $10^{-3}$ for 2,000 epochs. For all instances, the MSEs on the validation sets indicate that the models have successfully converged.

\begin{figure}[t]
\centering
\begin{minipage}{0.6\textwidth}
    \centering
    \includegraphics[width=\textwidth]{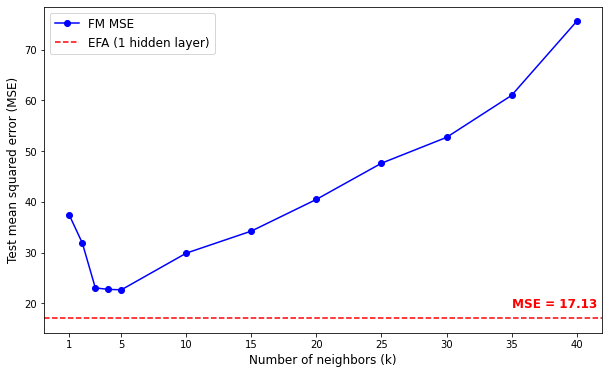}
    
\end{minipage}%
\hspace{0.03\textwidth} % Adjust the gap size here
\begin{minipage}{0.35\textwidth}
    \centering
    \begin{tabular}{c c}
    \hline
    EFE variant & Test MSE \\
    \hline
    $\ell = 1$ & 17.135 \\
    \hline
    $\ell = 2$ & 11.522 \\
    \hline
    $\ell = 4$ & 11.224 \\
    \hline
    $\ell = 4$, 1D ahead & \textbf{9.248} \\
    \hline
    $\ell = 4$, 2D ahead & 10.420 \\
    \hline
    \end{tabular}
\end{minipage}
\caption{EFA outperforms FM across all neighbor sizes ($k$), even with $\ell = 1$ hidden layer. The performance of EFA can be further enhanced by increasing the number of hidden layers ($\ell$) and expanding the context to also include temperatures of all cities---including itself---from the previous day (\textit{1D ahead}) or the previous two days (\textit{2D ahead}).
}
\label{fig:fm-efa-exp-1}
\end{figure}

\noindent
\textbf{Results.} Figure \ref{fig:fm-efa-exp-1} summarizes the test MSEs for FM across various values of $k$, where $k$ denotes the number of other time series included in the summation. From the results, it is evident that $k = 5$ yields the lowest test MSE (22.685) among all FM configurations, highlighting a balance between including useful information and excluding irrelevant or noisy signals. On the other hand, EFA---which considers all other time series and utilizes the self-attention mechanism to learn the context set in a data-driven manner---produces a significantly lower test MSE (17.135), even with a single self-attention layer.

We can also observe from Figure \ref{fig:fm-efa-exp-1} that adding more hidden layers in the EFA model results in further improvements in model performance. Similarly, expanding the context to include temperatures from previous days contributes to enhanced performance. To achieve this, we incorporate a learned 32-dimensional embedding that encodes the time step of each observation.

% !TeX root = ./main.tex

\section{Discussion}
\label{sec:disc}
This paper presents exponential family attention (EFA), a class of probabilistic models that extends self-attention \citep{vaswani2017attention,devlin2019bert} to handle both sequential and non-sequential data beyond text. By leveraging self-attention, EFA captures complex, non-linear interactions among elements in the context, integrates order information to model sequential dependencies, and adopts a data-driven approach to defining context dependencies. We showcase three real-world applications of EFA: spatiotemporal time series, market baskets, and movie ratings. We demonstrate that in each scenario, EFA subsumes a variant of linear latent factor models as special cases while overcoming some critical limitations in modeling complex sequential data. Theoretically, we establish an identifiability result along with an excess loss generalization guarantee for EFA. Finally, we evaluate EFA's empirical performance through synthetic and real data experiments, demonstrating significant improvements over existing baselines. 

% For future work, extending EFA to handle multimodal data with hierarchical context structures can be a promising avenue.

\noindent
\textbf{Acknowledgements.} This research was supported in part by the Office of Naval Research under grant number N00014-23-1-2590, the National Science Foundation under Grant No. 2231174, No. 2310831, No. 2428059, and a Michigan Institute for Data Science Propelling Original Data Science (PODS) grant.

\newpage
\bibliographystyle{chicago.bst}
\bibliography{biblio}

\newpage
\appendix
\bigskip
\begin{center}
{\large\bf SUPPLEMENTARY MATERIAL}
\end{center}

% !TeX root = ./jasa-attention.tex

\section{Related work} 

\label{sec:related}
Our work builds upon key concepts from latent factor models, exponential family embeddings, and the self-attention mechanism.

\noindent
\textbf{Latent variable models.} Latent factor models have been widely used and studied for their ability to analyze high-dimensional data and uncover hidden structures. \citet{sun2016latent} proposed an $L_1$-regularized method for selecting latent variables in multidimensional item response theory models. \citet{chen2019joint} studied theoretical properties of the joint maximum likelihood estimator for exploratory item factor analysis. \citet{chen2020structured} analyzed structured latent factor models, focusing on how design constraints affect their identifiability and estimation.

Recently, \citet{gu2023joint} proposed a joint maximum likelihood (JML) method for structured latent attribute models and studied its statistical properties. 
 \citet{gu2023bayesian} developed a new class of multilayer latent structure models for discrete data. \citet{ouyang2024statistical} developed covariate-adjusted generalized factor models and proposed a JML estimation method. \citet{lyu2024degree} introduced degree-heterogeneous latent class models that can capture nested within-class quantitative heterogeneity.

\noindent
\textbf{Exponential family embeddings.} Some works on latent variable models fall under the umbrella of exponential family embeddings (EFE) \citep{rudolph2016exponential}---an extension of 
continuous bag of words \citep{mikolov2013distributed} to model exponential family data. \citet{liu2017zero} proposed zero-inflated EFE, incorporating an exposure indicator to account for unrelated zero observations and improving the embeddings' ability to represent item relationships. \citet{rudolph2017structured} extended EFE by introducing hierarchical modeling and amortization, resulting in embeddings that adapt across related groups of data. \citet{rudolph2018dynamic} developed dynamic embeddings to capture how word meanings evolve over time.

EFE has also been adapted to graph-based and high-dimensional data. \citet{celikkanat2020exponential} modified EFE, building on the skip-gram model by \citet{mikolov2013distributed} to capture interaction patterns between nodes in random walks. \citet{lu2022nonparametric} employed nonparametric priors, allowing nodes to have multiple latent representations for greater flexibility in graph embeddings. \citet{jagtap2021multiomics} tailored EFE to model interactions among biological entities in high-dimensional omics data, leveraging exponential family distributions to learn diverse omics modalities.

Finally, EFE has inspired theoretical advancements and applications in various domains. \citet{krstovski2018equation} extended EFE to model semantic representations of mathematical equations, introducing separate embeddings for equations and words. \citet{yu2018provable} established foundational results for EFE, showing that embedding structures can be learned with limited observations under various data structures. \citet{lin2021exponential} developed a non-linear embedding method based on alternating minimization for single-cell RNA-seq data and proved its statistical consistency. 
%In contrast to these works, our work improves EFE by employing the self-attention mechanism \citep{vaswani2017attention}.

\noindent
\textbf{Self-attention mechanism.} First introduced in the Transformer model \citep{vaswani2017attention}, self-attention marked a significant departure from traditional recurrent and convolutional neural networks in language modeling. By assigning attention weights to every pair of input elements, self-attention allows models to capture dependencies across all elements, regardless of their distance in the sequence. This dynamic mechanism facilitates the modeling of interactions that are both flexible and data-driven. Additionally, self-attention incorporates positional embeddings, enabling the model to process sequential data without relying on recurrent structures.

Self-attention has served as the backbone of many modern large language models, including BERT \citep{devlin2019bert}, RoBERTa \citep{liu2019roberta}, and GPT-4 \citep{achiam2023gpt}, which have achieved remarkable performance on natural language processing benchmarks. Despite its success, self-attention is not inherently designed to handle high-dimensional sequence or spatial data in other domains, such as movie ratings, market baskets, and time series. Our work addresses this gap by introducing exponential family attention (EFA), a probabilistic method based on self-attention that extends its applicability to exponential family data beyond text domains.

\newpage
\section{Linear identifiability in \citet{roeder2021linear}}
\label{app:iden}
Suppose we are given a data set of the form $p_{\mathcal{D}}(x,y,S)$, where $x$, $y$ and $S$ represent the input variable, the target variable, and the set of all possible values of $y$ given $x$, respectively. Moreover, let $\Theta$ denote the set of all possible parameter values. 

Let $p_{\bm{\theta}} := p_{\bm{\theta}}(y \mid x, S)$. Traditionally, identifiability is said to hold if for any $\bm{\theta}', \bm{\theta}^* \in \Theta$, $p_{\bm{\theta}'} = p_{\bm{\theta}^*}$ implies $\bm{\theta}' = \bm{\theta}^*$. However, when $p_{\bm{\theta}}$ is parameterized by a neural network, the usual notion of identifiability is unlikely to be satisfied due to overparameterization. \citet{roeder2021linear} defines a more realistic version of identifiability (called \textit{linear identifiability}) for neural network-based models of the form
\begin{equation}
\label{eq:nn}
    p_{\bm{\theta}}(y \mid x, S) = \frac{\exp \left( f_{\bm{\theta}}(x)^\top g_{\bm{\theta}}(y) \right)}{\sum_{y' \in S} \exp \left( f_{\bm{\theta}}(x)^\top g_{\bm{\theta}}(y') \right)},    
\end{equation}
where $f_{\bm{\theta}}$ and $g_{\bm{\theta}}$ are neural networks such that $f_{\bm{\theta}}(\cdot), g_{\bm{\theta}}(\cdot) \in \mathbb{R}^M$. Note that the model family in Equation \eqref{eq:nn} comprises numerous well-known models, including the unidirectional and bidirectional variants of BERT \citep{devlin2019bert}. Identifiability is then defined \textit{up to a linear transformation}, as given by Definition \ref{def:iden}.

\begin{definition}[Linear identifiability \citep{roeder2021linear}]
\label{def:iden}
    Linear identifiability occurs if for any $\bm{\theta}', \bm{\theta}^* \in \Theta$, $p_{\bm{\theta}'} = p_{\bm{\theta}^*}$ implies $f_{\bm{\theta}'}(x) = \bm{A}f_{\bm{\theta}^*}(x)$ and $g_{\bm{\theta}'}(y) = \bm{B}g_{\bm{\theta}^*}(y)$ for some invertible $\bm{A}, \bm{B} \in \mathbb{R}^{M \times M}$.
\end{definition}

\citet{roeder2021linear} establishes that the model family in Equation \eqref{eq:nn} is linearly identifiable if the following \textit{diversity assumptions} are met.

\begin{assumption}[Diversity I]
\label{asm:div-y}
    For any $\bm{\theta}', \bm{\theta}^*, x$ such that $p_{\bm{\theta}'} = p_{\bm{\theta}^*}$, we can construct a set of distinct tuples $\{(y^{(i)}_{A}, y^{(i)}_{B})\}_{i=1}^M$ by sampling $S \sim p_{\mathcal{D}}(S \mid x)$ and picking $y_{A}, y_{B} \in S$ such that the matrices $\bm{L}'$ and $\bm{L}^*$ are invertible. Here, the columns of $\bm{L}'$ and $\bm{L}^*$ are $(g_{\bm{\theta}'}(y^{(i)}_{A}) - g_{\bm{\theta}'}(y^{(i)}_{B}))$'s and $(g_{\bm{\theta}^*}(y^{(i)}_{A}) - g_{\bm{\theta}^*}(y^{(i)}_{B}))$'s, where $i \in [M]$.
\end{assumption}

\begin{assumption}[Diversity II]
\label{asm:div-x}
    For any $\bm{\theta}', \bm{\theta}^*, y$ such that $p_{\bm{\theta}'} = p_{\bm{\theta}^*}$, we can construct distinct tuples $\{(x^{(i)}, S^{(i)})\}_{i=1}^{M+1}$ such that $p_{\mathcal{D}}(x^{(i)}, y, S^{(i)}) > 0$ and the matrices $\bm{M}'$ and $\bm{M}^*$ are invertible. Here, the columns of $\bm{M}'$ and $\bm{M}^*$ are 
    $$\left[ -\log \sum_{y' \in S^{(i)}} \exp \left( f_{\bm{\theta}'}(x^{(i)})^\top g_{\bm{\theta}'}(y') \right); f_{\bm{\theta}'}(x^{(i)}) \right]^\top$$ and
    $$\left[ -\log \sum_{y^* \in S^{(i)}} \exp \left( f_{\bm{\theta}^*}(x^{(i)})^\top g_{\bm{\theta}^*}(y^*) \right); f_{\bm{\theta}^*}(x^{(i)}) \right]^\top,$$
    where $i \in [M+1]$.
\end{assumption}

\begin{remark}[On the satisfiability of Assumptions \ref{asm:div-y} and \ref{asm:div-x}]
    For a $K$-class supervised classification problem, we need $M \leq K - 1$ for Assumption \ref{asm:div-y} to hold, as mentioned in Section 3.3 of \citet{roeder2021linear}. On the other hand, Assumption \ref{asm:div-x} typically holds as multiple $x$'s can be associated with a fixed $y$.
\end{remark}

\newpage
\section{Proof of Theorem \ref{thm:iden}}
\label{app:pf-iden}
\textit{Proof.} Note that by Theorem 1 of \citet{roeder2021linear}, the first condition in Definition \ref{def:lin-id-efa} holds under Assumptions \ref{asm:div-y-efa} and \ref{asm:div-x-efa}. We now show that under Assumptions \ref{asm:struc} and \ref{asm:div-add}, the second condition in Definition \ref{def:lin-id-efa} also holds.

Take any $\bm{\theta}_2', \bm{\theta}_2^* \in \Theta_2$ such that $p_{\bm{\theta}_2'}(y_i \mid x_{1:i}, y_{1:i-1}) = p_{\bm{\theta}_2^*}(y_i \mid x_{1:i}, y_{1:i-1})$. By Assumptions \ref{asm:struc} and \ref{asm:div-add}, we have $(\bm{N}')^\top L_{\bm{\theta}_2'} = (\bm{N}^*)^\top L_{\bm{\theta}_2^*}$, whence $L_{\bm{\theta}_2'} = (\bm{N}^* (\bm{N}')^{-1})^\top  L_{\bm{\theta}_2^*}$. Let $\bm{D} := (\bm{N}^* (\bm{N}')^{-1})^\top$. We then have $(\bm{N}')^\top = (\bm{N}^*)^\top \bm{D}^{-1}$, whence $\bm{N}' = (\bm{D}^{-1})^\top \bm{N}^*$. This implies $K_{\bm{\theta}_2'}(\underline{y}) = \bm{C} K_{\bm{\theta}_2^*}(\underline{y})$ where $\bm{C} := (\bm{D}^{-1})^\top$, completing the proof. $\blacksquare$

\newpage

\section{Proof of Theorem \ref{thm:gen-gua}}
\label{pf:theo}
\textit{Proof.} Define
$$X_{\bm{\theta}} := \frac{1}{F} \sum_{f=1}^F \ell(\bm{\theta}; Z^f) -  \mathbb{E}_{{Z}}\left[\ell(\bm{\theta}; Z)\right],$$
where $\{{Z}^f\}_{f \in [F]}$ are i.i.d. copies of ${Z} \sim P, P \sim \pi$. We now apply Proposition B.4 of \citet{bai2024transformers}. Before that, we need to check if the assumptions are satisfied.

First, note that Example 5.8 of \citet{wainwright2019high} yields $$\log N(\delta; B_{\vertiii{\cdot}}(r), \vertiii{\cdot}) \leq (L(3MD^2 + 2DD') + K'(K + \tau))\log(1 + 2r/\delta),$$
where $B_{\vertiii{\cdot}}(r)$ is any ball of radius $r$ under the norm $\vertiii{\cdot}$. 

Moreover, it is easy to see that
\begin{align*}
    \left| \ell(\bm{\theta}; {Z}) \right| &= \left| \frac{1}{2I} \sum_{i=1}^I \left( y_i - \textrm{clip}_{B_y}(\textrm{read}_y(\textrm{TF}_{\bm{\theta}}^R(\textrm{Exp}_{\bm{\theta}}^R(Z_i)))) \right)^2 \right| \\
    &\leq 2B_y^2
\end{align*}
using the elementary inequality $(a-b)^2 \leq 2(a^2+b^2)$, whence $\ell(\bm{\theta}; {Z})$ is $2B_y^2$-sub-Gaussian.

Finally, observe that the triangle inequality gives us
$$\left| \ell(\bm{\theta}; {Z}) - \ell(\bm{\tilde{\theta}}; {Z})\right| \leq \frac{1}{2I} \sum_{i=1}^I 4B_y \cdot \vertii{\textrm{TF}_{\bm{\theta}}^R(\textrm{Exp}_{\bm{\theta}}^R({Z}_{i})) - \textrm{TF}_{\bm{\tilde{\theta}}}^R(\textrm{Exp}_{\bm{\tilde{\theta}}}^R({Z}_{i}))}_{2, \infty}.$$
For any $Z$ such that the $\vertii{\cdot}_{2, \infty}$ norm of the first $\tau$ rows is at most $R$, we can apply Corollary L.1 of \citet{bai2024transformers}, the fact that $\vertii{\textrm{Exp}_{\bm{\theta}}^R({Z}) - \textrm{Exp}_{\bm{\tilde{\theta}}}^R({Z})}_{2, \infty} \leq BR \cdot \vertiii{\bm{\theta} - \bm{\tilde{\theta}}}$, and the triangle inequality to obtain
\begin{align*}
\vertii{\textrm{TF}_{\bm{\theta}}^R(\textrm{Exp}_{\bm{\theta}}^R({Z})) - \textrm{TF}_{\bm{\tilde{\theta}}}^R(\textrm{Exp}_{\bm{\tilde{\theta}}}^R({Z}))}_{2, \infty} \leq (L+1)B_H^{L}B_{\Theta} \cdot \vertiii{\bm{\theta} - \bm{\tilde{\theta}}},
\end{align*}
where $B_H := (1+B^2)(1+B^2R^3)$ and $B_{\Theta} := BR(1+BR^2+B^3R^2)$.
This implies that $$\left| \ell(\bm{\theta}; {Z}) - \ell(\bm{\tilde{\theta}}; {Z})\right| \leq 2B_y (L+1)B_H^{L}B_{\Theta} \cdot \vertiii{\bm{\theta} - \bm{\tilde{\theta}}},$$ i.e., $\ell(\bm{\theta}; {Z}) - \ell(\bm{\tilde{\theta}}; {Z})$ is $2B_y (L+1)B_H^{L}B_{\Theta}$-sub-Gaussian.

Since all the assumptions of Proposition B.4 of \citet{bai2024transformers} are satisfied, we have
$$\sup_{\bm{\theta}} \left| X_{\bm{\theta}} \right| \leq 2C B_y^2 \sqrt{\frac{(L(3MD^2 + 2DD') + K'(K + \tau)) \log \left( 4 + \frac{4(L+1)B_H^L B_{\Theta}B}{B_y}\right) + \log \left(\frac{1}{\xi}\right) }{F}}$$
with a probability of at least $1 - \xi$. Theorem \ref{thm:gen-gua} thus follows upon realizing that
$$\log \left( 4 + \frac{4(L+1)B_H^L B_{\Theta}B}{B_y}\right) \leq 43L \cdot \log \left( 2 + \max \left( B, R, \frac{1}{2B_y} \right)\right)$$
and $L(\hat{\bm{\theta}}) \leq \inf_{\bm{\theta}} L(\bm{\theta}) + 2\sup_{\bm{\theta}} \left| X_{\bm{\theta}} \right|.$ $\blacksquare$

\newpage

\section{Proof of Proposition \ref{prop:bert-cbow}}
\label{app:pf-cbow}
\noindent
\textit{Proof.} Set $\bm{p}_i = \mathbb{0}_K$ for all $i \in [I]$, $\bm{\beta}_{\textrm{[MASK]}} = \mathbb{0}_K$, $\bm{W}^\mathcal{Q} = \bm{W}^\mathcal{K} = \mathbb{0}_{K \times K}$, and $\bm{W}^\mathcal{V} = \mathbb{I}_{K \times K}$. It is easy to see that
$X_{ii}' = \frac{1}{I} \left( \bm{\beta}_{x_1} + \dots + \bm{\beta}_{x_{i-1}} + \bm{\beta}_{x_{i+1}} + \dots + \bm{\beta}_{x_{I}}\right).$ From here, setting $\bm{\delta}_\ell = \frac{I}{I-1} \bm{\rho}_\ell$ and $\bm{\beta}_\ell = \bm{\alpha}_\ell$ for each $\ell \in [D]$ reduces the bidirectional EFA model to the claimed linear latent factor model, completing the proof. $\blacksquare$

\newpage

\section{Proof of Proposition \ref{prop:gauss}}
\label{app:pf-gauss}
\noindent
\textit{Proof.} Set $\bm{\lambda_1}(y) = y$ when $y \neq \texttt{[MASK]}$ and $\bm{\lambda_1}(\texttt{[MASK]}) = 0$, $\bm{\bar{W}}^\mathcal{V} = \textrm{diag}([0, \dots, 0, 1]) \in \mathbb{R}^{(K + 1) \times (K+1)}$, $\bm{\bar{W}}^\mathcal{Q} = \bm{\bar{W}}^\mathcal{K} = \textrm{diag}([1, \dots, 1, 0]) \in \mathbb{R}^{(K + 1) \times (K+1)}$, $\bm{g} = \bm{h}$, and $\bm{\lambda_2}(z) = (K+1)z_{K+1}$, where $z_{K+1}$ denotes the $(K+1)$-th element of the vector $z$. Observe that
$$\bm{\lambda_2}(Y'_{ii}) = \bm{h}(\tau_i))^\top \sum_{j \in [I], j \neq i} \bm{h}(\tau_j) y_j,$$
completing the proof. $\blacksquare$

\newpage

\section{Proof of Proposition \ref{prop:rating}}
\label{app:pf-rating}
\noindent
\textit{Proof.} To demonstrate the equivalence, we set the following parameters: 
$\bm{\bar{\beta}} = \bm{\rho}$, $\bm{\bar{\delta}} = \frac{1}{I-1}\bm{\alpha}$, $\bm{\lambda_1}(y) = y$ when $y \neq \texttt{[MASK]}$ and $\bm{\lambda_1}(\texttt{[MASK]}) = 0$, $\bm{\bar{p}}_i = \mathbb{0}_{2K+1}$ for all $i \in [I]$, $\bm{\bar{W}}^\mathcal{V} = \textrm{diag}([0, \dots, 0, 1]) \in \mathbb{R}^{(2K + 1) \times (2K+1)}$, $\bm{\bar{W}}^\mathcal{Q} = \textrm{diag}([1, \dots, 1, 0, \dots, 0]) \in \mathbb{R}^{(2K + 1) \times (2K+1)}$ with $K$ $1$'s and $K+1$ $0$'s, 
$$\bm{\bar{W}}^\mathcal{V} = \begin{bmatrix}
\mathbf{0}_{K \times K} & \mathbf{0}_{K \times K} & \mathbf{0}_K \\ \hline
\mathbb{I}_{K \times K} & \mathbf{0}_{K \times K} & \mathbf{0}_K \\ \hline
\mathbf{0}_K^\top & \mathbf{0}_K^\top & 0
\end{bmatrix} \in \mathbb{R}^{(2K + 1) \times (2K+1)},$$
and $\bm{\lambda_2}(z) = (2K+1)z_{2K+1}$, where $z_{2K+1}$ denotes the $(2K+1)$-th element of the vector $z$. With these settings, we have
$$\bm{\lambda_2}(Y'_{ii}) = \frac{1}{I-1} \rho_{x_i}^\top \sum_{j \in [I], j \neq i} \alpha_{x_j} y_j,$$
demonstrating that this specific instance of the EFA model is equivalent to a linear latent factor model. This completes the proof. $\blacksquare$

\newpage
\section{Experimental details}
\label{app:ex}

\subsection{Synthetic data (Section \ref{sec:synth})}

\label{app:synth-details}

\noindent
\textbf{Data generating process.} The data generation process is as follows:
\begin{enumerate}[leftmargin=0.6cm]
    \item There are 5 different movies (numbered 1 to 5). Each user rates each movie once, but the order in which the movies are rated is randomly determined, with each of the 120 possible permutations being equally likely.
    \item By default, each movie's rating is independently drawn from a $\mathcal{N}(3,1)$ distribution.
    \item However, there are a few special cases that modify this default rating:
    \begin{enumerate}[leftmargin=0.6cm]
        \item If movie 2 is rated after movie 1 (not necessarily directly), movie 2's rating follows a $\mathcal{N}(1,1)$ distribution. Otherwise, movie 2's rating follows a $\mathcal{N}(5,1)$ distribution.
        \item If movie 4 is rated \textit{directly} after movie 3, movie 4's rating follows a $\mathcal{N}(1,1)$ distribution. Similarly, if movie 3 is rated \textit{directly} after movie 4, movie 3's rating follows a $\mathcal{N}(1,1)$ distribution.
        \item If movie 5 is the last one rated, its rating follows a $\mathcal{N}(5,1)$ distribution.
    \end{enumerate}
\end{enumerate}

In real life, the scenarios in cases (a) to (c) can indeed occur. In case (a), consider similarly themed movies like \textit{The Hunger Games} and \textit{Divergent}. Watching the second movie after the first might feel less engaging because the plots are too similar. However, watching them in reverse order can make the first movie feel fresher and more exciting, resulting in a higher rating. In case (b), think of back-to-back episodes in \textit{The Lord of the Rings}. Watching one episode immediately after the other may feel like binge-watching, which can reduce enjoyment. Alternatively, watching one episode without the proper buildup from the previous one could make it feel out of place. In case (c), movies like \textit{Avengers: Endgame} are designed to be watched last for maximum impact, and watching them earlier can diminish their emotional payoff.

\noindent
\textbf{Model configurations.} For FM, we follow the model in Proposition \ref{prop:rating}, except that we remove the $\exp(\cdot)$ layers and replace the Possion distribution with the Gaussian distribution with a known variance $\sigma^2$. That is,
$$y_i  \mid y_{-i} \sim \mathcal{N}\left(\frac{1}{I-1} \bm{\rho}_{x_i}^\top \sum_{j \in [I], j \neq i} \bm{\alpha}_{x_j} y_j, \sigma^2 \right)$$ for bidirectional and
$$y_i  \mid y_{<i} \sim \mathcal{N}\left( \frac{1}{I-1} \bm{\rho}_{x_i}^\top \sum_{j \in [I], j < i} \bm{\alpha}_{x_j} y_j, \sigma^2 \right)$$ for unidirectional. As mentioned in Section \ref{sec:arch}, maximizing the Gaussian log-likelihood is equivalent to minimizing the mean squared error that does not depend on the choice of $\sigma^2$.

For EFA, to predict the rating of the $i$-th movie, we first present our data in the following matrix form:
\[
    Y_i := \left[
      \begin{array}{ccccccc}
        \bm{\bar{\beta}}_{x_1} & \ldots & \bm{\bar{\beta}}_{x_{i-1}} & \bm{\bar{\beta}}_{x_i} & \bm{\bar{\beta}}_{x_{i+1}} & \ldots & \bm{\bar{\beta}}_{x_{I}} \\
        \bm{\lambda_1}(y_1) & \ldots & \bm{\lambda_1}(y_{i-1}) & \bm{\lambda_1}(\texttt{[MASK]}) & \bm{\lambda_1}(y_{i+1}) & \ldots & \bm{\lambda_1}(y_{I})
      \end{array}
    \right],
\]
where $\bm{\bar{\beta}}_\cdot$ and $\bm{\lambda_1}(\cdot)$ represent 32-dimensional categorical embeddings for movies and ratings, respectively, with a separate embedding for the \texttt{[MASK]} rating. We then apply 2 consecutive self-attention layers to $Y_i$, where each layer consists of 2 heads with residual connections, resulting in $Y_i'$. The column of $Y_i'$ corresponding to the masked token (i.e., $Y_{ii}'$) is then passed through $\bm{\lambda_2}(\cdot)$, where $\bm{\lambda_2}(\cdot)$ consists of a 32-dimensional ReLU dense layer, followed by a 1-dimensional linear dense layer.

\noindent
\textbf{Optimization.} We maximize the (pseudo) log-likelihood utlizing the Adam optimizer \citep{kingma2014adam} with a learning rate of $10^{-4}$ for up to 1,000 epochs. We apply early stopping on the validation set, with a patience of 10 epochs.

\subsection{MovieLens ratings (Section \ref{sec:movie})}
\label{app:movielens}
In Experiment 1, for FM, we follow the model in Proposition \ref{prop:bert-cbow}, i.e.,
$$x_i \mid x_{-i} \sim \textrm{Categorical}\left( \sigma \left( \frac{1}{I-1} \bm{\rho}^\top \sum_{j \in [I], j \neq i} \bm{\alpha}_{x_j} \right) \right)$$
for bidirectional and
$$x_i \mid x_{<i} \sim \textrm{Categorical}\left( \sigma \left( \frac{1}{I-1} \bm{\rho}^\top \sum_{j \in [I], j < i} \bm{\alpha}_{x_j} \right) \right)$$ for unidirectional. For EFA, we follow the model in Section \ref{sec:self-attn-lang}, but $X_i'$ is derived via applying 2 consecutive self-attention layers to $X_i$, where each layer consists of 2 heads with residual connections.

In Experiment 2, for FM, we follow the model in Proposition \ref{prop:rating}, i.e.,
$$y_i - 1 \mid y_{-i} \sim \textrm{Poisson}\left( \exp\left(\frac{1}{I-1} \bm{\rho}_{x_i}^\top \sum_{j \in [I], j \neq i} \bm{\alpha}_{x_j} y_j \right)\right);$$
$$y_i  \mid y_{-i} \sim \textrm{Poisson}\left( 1 + \exp\left(\frac{1}{I-1} \bm{\rho}_{x_i}^\top \sum_{j \in [I], j \neq i} \bm{\alpha}_{x_j} y_j \right)\right)$$ for bidirectional and
$$y_i - 1 \mid y_{<i} \sim \textrm{Poisson}\left( \exp\left(\frac{1}{I-1} \bm{\rho}_{x_i}^\top \sum_{j \in [I], j < i} \bm{\alpha}_{x_j} y_j \right)\right);$$
$$y_i  \mid y_{<i} \sim \textrm{Poisson}\left( 1 + \exp\left(\frac{1}{I-1} \bm{\rho}_{x_i}^\top \sum_{j \in [I], j < i} \bm{\alpha}_{x_j} y_j \right)\right)$$ for unidirectional.

For EFA, to predict the rating of the $i$-th movie, we first present our data in the following matrix form:
\[
    Y_i := \left[
      \begin{array}{ccccccc}
        \bm{\bar{\beta}}_{x_1} & \ldots & \bm{\bar{\beta}}_{x_{i-1}} & \bm{\bar{\beta}}_{x_i} & \bm{\bar{\beta}}_{x_{i+1}} & \ldots & \bm{\bar{\beta}}_{x_{I}} \\
        \bm{\lambda_1}(y_1) & \ldots & \bm{\lambda_1}(y_{i-1}) & \bm{\lambda_1}(\texttt{[MASK]}) & \bm{\lambda_1}(y_{i+1}) & \ldots & \bm{\lambda_1}(y_{I})
      \end{array}
    \right],
\]
where $\bm{\bar{\beta}}_\cdot$ and $\bm{\lambda_1}(\cdot)$ represent 32-dimensional categorical embeddings for movies and ratings, respectively, with a separate embedding for the \texttt{[MASK]} rating. We then apply 2 consecutive self-attention layers to $Y_i$, where each layer consists of 2 heads with residual connections, resulting in $Y_i'$. The column of $Y_i'$ corresponding to the masked token (i.e., $Y_{ii}'$) is then passed through $\exp(\bm{\lambda_2}(\cdot))$, where $\bm{\lambda_2}(\cdot)$ consists of a 32-dimensional ReLU dense layer, followed by a 1-dimensional linear dense layer.

\subsection{Instacart baskets (Section \ref{sec:groc})}
For FM, we follow the model in Proposition \ref{prop:bert-cbow}, i.e.,
$$x_i \mid x_{-i} \sim \textrm{Categorical}\left( \sigma \left( \frac{1}{I-1} \bm{\rho}^\top \sum_{j \in [I], j \neq i} \bm{\alpha}_{x_j} \right) \right)$$
for bidirectional and
$$x_i \mid x_{<i} \sim \textrm{Categorical}\left( \sigma \left( \frac{1}{I-1} \bm{\rho}^\top \sum_{j \in [I], j < i} \bm{\alpha}_{x_j} \right) \right)$$ for unidirectional.

For EFA, we follow the model in Section \ref{sec:self-attn-lang}, but $X_i'$ is derived via applying 2 consecutive self-attention layers to $X_i$, where each layer consists of 2 heads with residual connections.

\subsection{US cities temperature (Section \ref{sec:temp})}
For FM, each 2-dimensional coordinate $\tau_i$ is first passed through a 128-dimensional dense layer with ReLU activation, followed by a 32-dimensional dense layer with a linear activation, resulting in $\bm{h}(\tau_i)$. The model follows Proposition \ref{prop:gauss}, i.e., 
$$y_i \mid y_{-i} \sim \mathcal{N}\left( (\bm{h}(\tau_i))^\top \sum_{j \in [I], j \neq i} \bm{h}(\tau_j) y_j, \sigma^2 \right).$$

For EFA, to predict the temperature of the $i$-th city, we first organize our data in the following matrix form:
\[
\left[
  \begin{array}{ccccccc}
    \bm{g}(\tau_1) & \ldots & \bm{g}(\tau_{i-1}) & \bm{g}(\tau_i) & \bm{g}(\tau_{i+1}) & \ldots & \bm{g}(\tau_I) \\
    \bm{\lambda_1}(y_1) & \ldots & \bm{\lambda_1}(y_{i-1}) & \bm{\lambda_1}(\texttt{[MASK]}) & \bm{\lambda_1}(y_{i+1}) & \ldots & \bm{\lambda_1}(y_{I}) \\
  \end{array}
\right],
\]
where $\bm{g}(\cdot)$ and $\bm{\lambda_1}(\cdot)$ both represent a 128-dimensional dense layer with ReLU activation, followed by a 32-dimensional dense layer. We have a separate embedding for the \texttt{[MASK]} temperature. Each column of the matrix is then passed through a 64-dimensional ReLU dense layer and a 32-dimensional linear dense layer, resulting in $Y_i$. We then apply 4 consecutive self-attention layers to $Y_i$, where each layer consists of 2 heads with residual connections, a 64-dimensional ReLU dense layer and layer normalization, resulting in $Y_i'$. The column of $Y_i'$ corresponding to the masked token (i.e., $Y_{ii}'$) is then passed through $\bm{\lambda_2}(\cdot)$, which consists of 128-dimensional and 16-dimensional ReLU dense layers, followed by a 1-dimensional linear dense layer.

To also include temperatures from the previous day, we simply modify the input matrix as follows:
\[
\left[
  \begin{array}{cccccccccc}
    \bm{g}(\tau_1) & \ldots & \bm{g}(\tau_{i-1}) & \bm{g}(\tau_i) & \bm{g}(\tau_{i+1}) & \ldots & \bm{g}(\tau_I) & \bm{g}(\tau_1) & \ldots & \bm{g}(\tau_I) \\
    \bm{\lambda_d}(0) & \ldots & \bm{\lambda_d}(0) & \bm{\lambda_d}(0) & \bm{\lambda_d}(0) & \ldots & \bm{\lambda_d}(0) & \bm{\lambda_d}(1) & \ldots & \bm{\lambda_d}(1)\\
    \bm{\lambda_1}(y_1) & \ldots & \bm{\lambda_1}(y_{i-1}) & \bm{\lambda_1}(\texttt{[MASK]}) & \bm{\lambda_1}(y_{i+1}) & \ldots & \bm{\lambda_1}(y_{I}) & \bm{\lambda_1}(\bar{y}_1) & \ldots & \bm{\lambda_1}(\bar{y}_I)\\
  \end{array}
\right],
\]
where $\bm{\lambda_d}(\cdot)$ is a learned 32-dimensional embedding for the day indicator and $\bar{y}_i$ is the temperature of city $i$ on the previous day. When including temperatures from the past few days, the same logic follows.

\newpage

\end{document}